\documentclass[pdflatex,sn-mathphys-ay]{sn-jnl}

\usepackage{amssymb}
\usepackage{amsmath}
\usepackage{array}
\usepackage{graphicx}
\usepackage{longtable}
\usepackage{float}
\usepackage{multirow}
\usepackage{makecell}
\usepackage{booktabs}
\usepackage{tabularx}
\usepackage{pdflscape}

\begin{document}

\title[AI4Radiology]{Representation Paradigms in AI-based 3D Radiological Image Reconstruction: A Systematic Review}

\author[1,2,3]{\fnm{Yuezhe} \sur{Yang}}\email{wa2214014@stu.ahu.edu.cn}

\author[3]{\fnm{Lei} \sur{Bi}}\email{lei.bi@sjtu.edu.cn}

\author[2]{\fnm{Boyu} \sur{Yang}}\email{wa2324289@stu.ahu.edu.cn}

\author[2]{\fnm{Yaqian} \sur{Wang}}\email{wa2314181@stu.ahu.edu.cn}

\author[2]{\fnm{Yang} \sur{He}}\email{wa2314093@stu.ahu.edu.cn}

\author[3]{\fnm{Yige} \sur{Peng}}\email{yige.peng@sjtu.edu.cn}

\author[2]{\fnm{Zhe} \sur{Jin}}\email{jinzhe@ahu.edu.cn}

\author*[2]{\fnm{Xingbo} \sur{Dong}}\email{xingbo.dong@ahu.edu.cn}

\author*[1]{\fnm{Jinman} \sur{Kim}}\email{jinman.kim@sydney.edu.au}

\affil[1]{\orgdiv{Biomedical Data Analysis and Visualisation (BDAV) Lab, School of Computer Science}, \orgname{The University of Sydney}, \orgaddress{\city{Sydney}, \country{Australia}}}

\affil[2]{\orgdiv{Anhui Provincial International Joint Research Center for Advanced Technology in Medical Imaging, School of Artificial Intelligence}, \orgname{Anhui University}, \orgaddress{\city{Hefei}, \state{Anhui}, \country{China}}}

\affil[3]{\orgname{Institute of Translational Medicine, Shanghai Jiao Tong University}, \orgaddress{\city{Shanghai}, \country{China}}}

\abstract{The demand for high-quality medical imaging in clinical practice and assisted diagnosis has made 3D image reconstruction in radiological imaging a key research focus. Artificial intelligence (AI) has emerged as a promising approach for improving reconstruction accuracy while reducing acquisition and processing time, thereby minimizing patient radiation exposure and discomfort and ultimately benefiting clinical diagnosis. This review surveys state-of-the-art AI-based 3D reconstruction algorithms in radiological imaging and organizes them into four representation families according to how the reconstructed target is parameterized: discrete grid representations, explicit basis expansion representations, explicit primitive representations, and implicit neural representations. In particular, the review clarifies the relationships among these representation forms and highlights radiance field methods as a specialized subtype of implicit neural representation. In addition, we summarize commonly used evaluation metrics and benchmark datasets for radiological image reconstruction. Finally, we discuss the current state of development, major challenges, and future research directions in this rapidly evolving field. Our project is available at: \url{https://github.com/Bean-Young/AI4Radiology}.}


\maketitle

\section{Introduction}
\label{sec1}

Radiology involves a series of imaging examinations used to visualize internal anatomical structures and physiological processes in the human body, including X-ray, magnetic resonance imaging (MRI), ultrasound, computed tomography (CT), and positron emission tomography (PET) \citep{sanghvi2010modalities}. In modern clinical practice, radiological imaging has become indispensable for early disease detection, diagnosis, treatment planning, and therapy monitoring \citep{shen2017deep}. To transform signals collected by imaging sensors into clinically interpretable images, reconstruction algorithms play a central role \citep{ben2021deep}. Among them, three-dimensional reconstruction is particularly important, because it provides volumetric representations of organs and tissues that help reveal infections, tumors, trauma, and abnormalities in blood vessels and internal organs \citep{singh20203d}.

More broadly, research on 3D reconstruction in radiological imaging lies at the intersection of applied mathematics, engineering, medicine, radiology, and computer science \citep{blackledge2005digital}. Its development has not followed a single linear path, but has instead been shaped by the interaction of imaging physics, inverse problem theory, machine learning, and computer graphics. In this work, we review the evolution of AI-based 3D reconstruction for radiological imaging from four closely related perspectives: traditional reconstruction algorithms \ref{1.1}, neural network-based methods \ref{1.2}, generative models \ref{1.3}, and representation paradigms \ref{1.4}. These perspectives should not be viewed as isolated branches. Rather, they have evolved through continuous interaction, with later methods often inheriting principles from earlier frameworks while introducing new forms of data modeling and representation.

\subsection{Development of Reconstruction Algorithms}
\label{1.1}

Early radiological reconstruction methods were primarily based on analytical or model-based formulations of the inverse problem \citep{gothwal2022computational}. These methods relied on explicit mathematical descriptions of the imaging system to recover images from measurements. Representative examples include Filtered Back Projection (FBP) for CT \citep{shi2015novel} and the Fast Fourier Transform (FFT) for MRI \citep{cooley1969fast,sumanaweera2005medical}. These approaches remain foundational because they are computationally efficient and can produce high-quality reconstructions when the data are noise-free and fully sampled \citep{jin2017deep}. However, such physics-driven methods mainly exploit the acquisition geometry and sampling characteristics of the imaging system \citep{fessler2017medical}, making them sensitive to noise, missing data, and modeling mismatch.

To improve robustness, iterative reconstruction methods were later introduced \citep{gordon1970algebraic}. Compared with analytical algorithms, iterative methods are more flexible in incorporating statistical models, physical constraints, and prior information, and therefore often provide better performance under noisy or incomplete measurements \citep{ravishankar2019image}. In emission tomography, one representative example is the Ordered Subset Expectation Maximization (OSEM) algorithm \citep{hudson1994accelerated}, which is an accelerated variant of Maximum Likelihood Expectation Maximization (MLEM) \citep{vardi1985statistical}. OSEM has been widely used in PET imaging \citep{yao2000performance,rapisarda2010image} because it improves convergence speed by dividing projection data into subsets and updating the image estimate iteratively. Nevertheless, traditional reconstruction algorithms still face limitations in modeling complex anatomical structures, heterogeneous noise patterns, and data variability in real clinical environments \citep{wang2020deep}.

\subsection{Advent of Neural Networks}
\label{1.2}

With the rapid development of artificial intelligence, neural networks introduced a new data-driven paradigm for radiological image reconstruction \citep{subasi2024artificial}. Their strong representation capacity and nonlinear modeling ability have greatly expanded the solution space of reconstruction algorithms, leading to increasing attention in 3D radiological imaging \citep{hosny2018artificial,wang2012machine}.

At an early stage, convolutional neural networks (CNNs) \citep{lecun1989backpropagation} demonstrated remarkable capability in hierarchical feature learning \citep{chen2025review}, which motivated their adoption in medical image reconstruction \citep{jurek2020cnn}. For example, Gong et al. incorporated a deep residual CNN into an iterative framework for PET reconstruction and improved image quality by leveraging inter-patient information \citep{gong2018iterative}. Despite their success, CNNs are inherently limited in modeling long-range dependencies and global context \citep{alzubaidi2021review}, which can hinder their performance in complex 3D reconstruction tasks.

Inspired by the Transformer architecture \citep{waswani2017attention}, Vision Transformers (ViT) \citep{dosovitskiy2020image} were introduced into medical imaging and rapidly became an important backbone for high-level image understanding \citep{azad2024advances}. Their global self-attention mechanism helps overcome some of the locality limitations of CNNs. However, ViT is primarily designed for feature extraction rather than inverse reconstruction, and its quadratic complexity can be problematic for high-resolution radiological images \citep{ali2023vision}. As a result, although neural networks have greatly advanced reconstruction quality, efficiency and scalability remain important concerns.

\subsection{Rise of Generative Models}
\label{1.3}

Another major development came from generative modeling. Instead of only learning deterministic mappings from input measurements to reconstructed images, generative models aim to capture the underlying data distribution and thus provide a stronger mechanism for prior modeling. Variational Autoencoders (VAE) \citep{kingma2013auto} introduced probabilistic latent representations, while Generative Adversarial Networks (GAN) \citep{goodfellow2014generative} employed adversarial training to generate realistic samples. Their strong generation capability soon attracted attention in radiological image reconstruction \citep{gothwal2022computational}. For instance, Luo et al. proposed a Transformer-GAN framework for reconstructing standard-dose PET images from low-dose PET acquisitions \citep{luo20213d}.

Although GAN-based methods often improve image sharpness and perceptual realism, their adversarial training process can be unstable and may introduce unnatural structures that are undesirable in medical applications \citep{sriram2024challenges}. More recently, diffusion models have emerged as a powerful alternative \citep{croitoru2023diffusion}. By learning a reverse denoising process from progressively corrupted data, diffusion models offer more stable training and flexible prior modeling. For example, AdaDiff uses an adaptive diffusion prior for accelerated MRI reconstruction and addresses domain shifts in both imaging operators and image distributions \citep{gungor2023adaptive}.

Nevertheless, generative models still face challenges in 3D medical imaging. In particular, they often struggle to ensure viewpoint consistency, preserve internal structural fidelity, and maintain computational efficiency when extended to volumetric reconstruction \citep{chen2024opportunities,molaei2023implicit}. These limitations motivate the search for more suitable representation paradigms for 3D radiological reconstruction.

\subsection{Emergence of Representation Paradigms}
\label{1.4}

In recent years, the focus of 3D reconstruction research has gradually shifted from network architecture alone to the question of how the reconstruction target itself should be represented. This representation-centered perspective is especially important in radiological imaging, because the choice of representation directly affects spatial continuity, memory efficiency, rendering speed, and the ability to integrate physical imaging models.

A major breakthrough in this direction came from implicit neural representations. In the natural image domain, Neural Radiance Fields (NeRF) demonstrated that a 3D scene can be modeled as a continuous neural field and rendered from arbitrary viewpoints with strong spatial consistency \citep{mildenhall2021nerf}. This idea was soon adapted to radiological imaging. For example, MedNeRF combined NeRF with GAN to reconstruct 3D-aware CT projections from single-view or few-view X-rays \citep{corona2022mednerf}, showing that continuous neural representations can reduce radiation exposure while preserving high-fidelity anatomical information.

At the same time, explicit primitive-based representations also advanced rapidly. While NeRF provides strong continuity and flexible rendering, its optimization and inference are often computationally expensive, which limits real-time clinical applicability \citep{wang2024neural}. In contrast, 3D Gaussian Splatting (3DGS) introduced an explicit primitive representation based on learnable Gaussian elements and achieved both high reconstruction fidelity and fast rendering \citep{kerbl20233d}. This idea has also inspired radiological applications. For instance, $R^2$-Gaussian is a Gaussian splatting framework for sparse-view tomographic reconstruction that improves both efficiency and reconstruction quality in volumetric imaging \citep{zha2024r}. More broadly, recent advances suggest that radiological image reconstruction can no longer be adequately described only in terms of architectures such as CNNs, Transformers, or GANs. Instead, it increasingly requires a representation-oriented taxonomy.

From this viewpoint, the methods reviewed in this survey can be organized into four representation families: discrete grid representations, explicit basis expansion representations, explicit primitive representations, and implicit neural representations. These four families provide a more precise and consistent description of how reconstructed targets are parameterized in modern radiological imaging systems.

\begin{figure}[htbp]
    \centering
    \includegraphics[width=\linewidth]{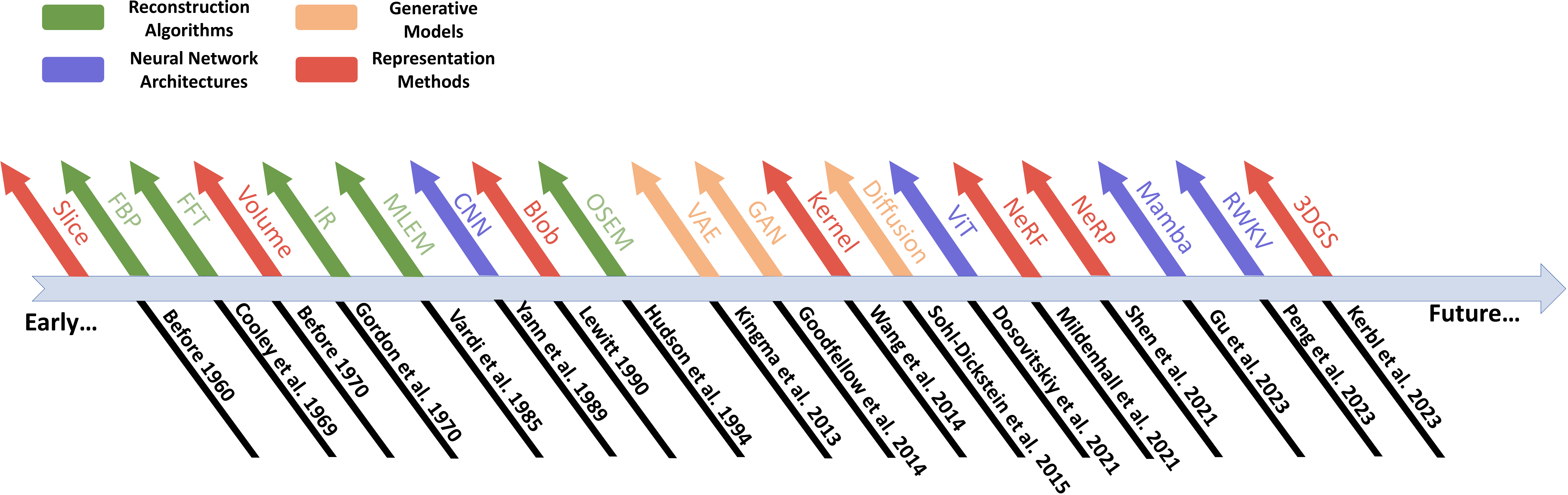} 
    \caption{A brief chronology of the development of techniques related to radiological image reconstruction in literature.}
    \label{fig:0}
\end{figure}

In summary, advances in artificial intelligence have profoundly transformed the field of 3D reconstruction in radiological imaging and have led to a rich variety of methods. The developmental roadmap is shown in Fig.\ref{fig:0}. Therefore, a systematic review is needed to clarify the relationships among these methods and to provide a structured reference for future research.

However, existing reviews often focus on specific model families or summarize only a subset of available technologies, frequently overlooking the emerging role of implicit imaging methods and representation-oriented analysis \citep{ahishakiye2021survey,gothwal2022computational,yasmin2012brain,xia2023physics}. This narrow focus makes it difficult for readers to fully understand how radiological imaging, inverse problems, and modern AI techniques have converged in recent years.

To address this gap, this review provides a comprehensive overview of AI-based 3D reconstruction in radiological imaging from a representation perspective. Specifically, we organize existing methods into four representation families: discrete grid representations, explicit basis expansion representations, explicit primitive representations, and implicit neural representations. This taxonomy enables a more consistent comparison of reconstruction strategies across CT, MRI, PET, SPECT, and 3D ultrasound. Based on literature published in recent years, we summarize the key methodological developments, benchmark datasets, evaluation metrics, current challenges, and future opportunities in the field. The primary contributions of this review are as follows:
\begin{itemize}
\item We provide a comprehensive review of AI-based 3D radiological image reconstruction across CT, MRI, PET, SPECT, US, and multi-modality scenarios.
\item We propose a representation-oriented taxonomy that organizes existing methods into four families: discrete grid, explicit basis expansion, explicit primitive, and implicit neural representations.
\item We present a refined overview of representative publicly available datasets and evaluation metrics for radiological image reconstruction.
\item We analyze current challenges and outline future research directions for AI-based 3D radiological image reconstruction.
\end{itemize}

The remainder of this survey is organized as follows. Section \ref{sec2} describes the literature review methodology. Section \ref{sec3} introduces publicly available datasets and evaluation metrics. Section \ref{sec4} presents discrete grid representations. Section \ref{sec5} introduces explicit basis expansion representations. Section \ref{sec6} summarizes explicit primitive representations. Section \ref{sec7} focuses on implicit neural representations. Section \ref{sec8} discusses the strengths and weaknesses of different representation families, as well as current challenges and future directions. Finally, Section \ref{sec9} concludes the review.

\begin{figure}[htbp]
    \centering
    \includegraphics[width=\linewidth]{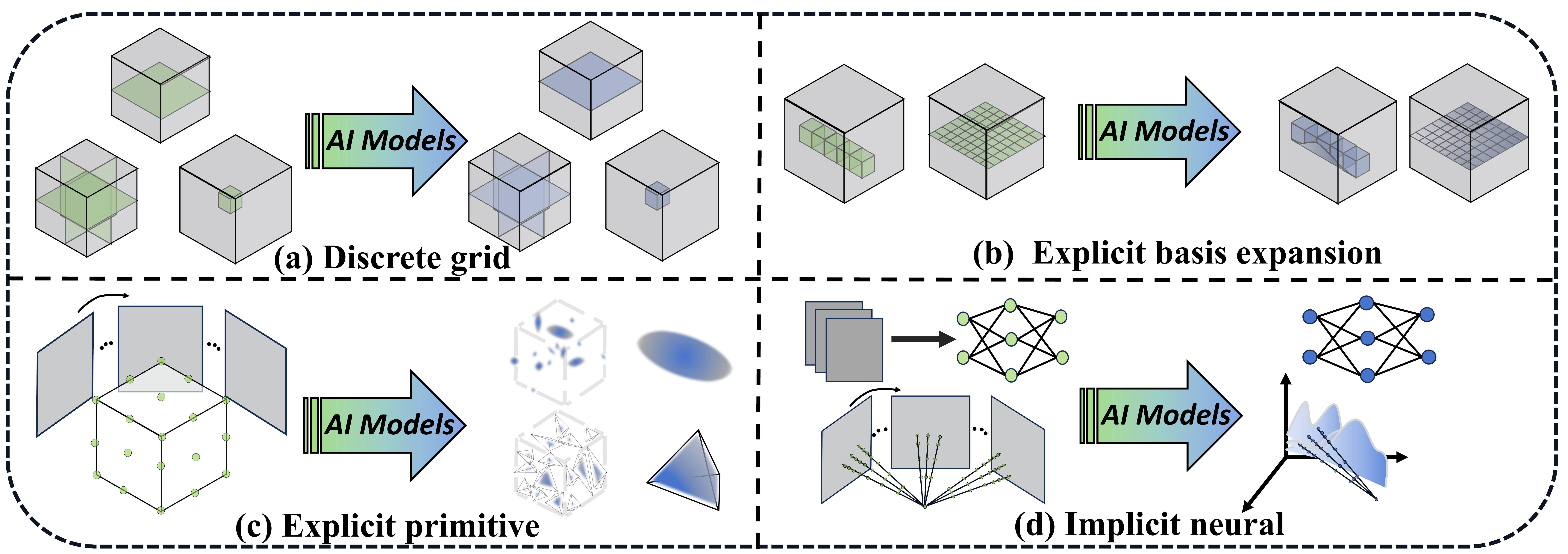} 
    \caption{Taxonomy of representation families for AI-based 3D reconstruction in radiological imaging.}
    \label{fig:1}
\end{figure}

\section{Methods}
\label{sec2}

\subsection{Literature Review}
The research on 3D reconstruction in radiological imaging and artificial intelligence primarily draws from the following databases: Web of Science \footnote{\url{https://www.webofscience.com/}}, Google Scholar \footnote{\url{https://scholar.google.com/}}, and Scopus \footnote{\url{https://www.scopus.com/}}. The search encompasses publications from the past five years.

\begin{figure}[htbp]
    \centering
    \includegraphics[width=0.6\textwidth]{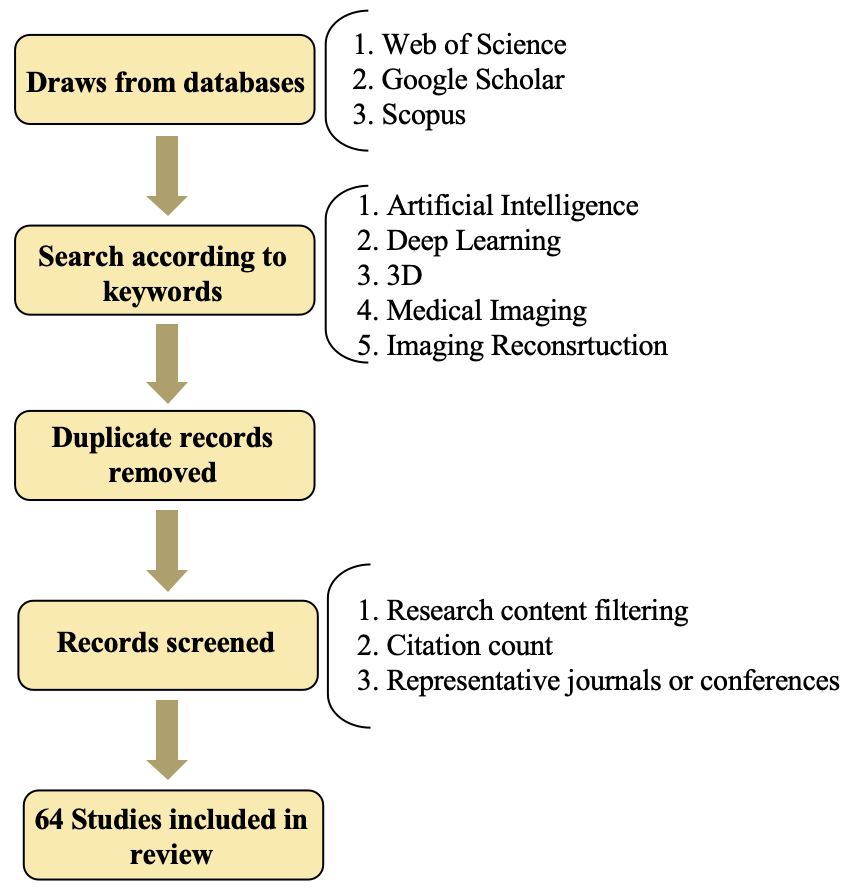} 
    \caption{Flowchart of the literature review.}
    \label{fig:2}
\end{figure}

Initially, the search results are refined using filtering criteria, such as article type and research field, to narrow the number of results. Next, duplicate records are excluded. Subsequently, the records are screened based on their titles, abstracts, and the study selection criteria. Studies that cannot be retrieved are then excluded. Finally, 64 studies that meet the inclusion criteria are incorporated into the systematic review. Fig.\ref{fig:2} illustrates the flowchart of the systematic literature review. The distribution of these studies by year is depicted in Fig.\ref{fig:3}\textbf{(a)}.

In Fig.\ref{fig:2}, our search keywords include “artificial intelligence,” “deep learning,” “3D,” “medical imaging,” and “imaging reconstruction.” Only research articles that contain all these keywords are included in our study. we apply three selection criteria: \textbf{1)} Research content filtering: Papers are selected based on their intrinsic research focus. We rigorously evaluate the relevance of each paper to ensure it is representative and relevant to our research scope. \textbf{2)} Citation count: We assess the citation count of each paper. Higher citation counts are prioritized, as a higher citation count often reflects greater recognition and impact within the field. \textbf{3)} Representative journals or conferences: Priority is given to papers published in prestigious journals or leading conferences, such as \textit{IEEE Transactions on Medical Imaging (T-MI)}, \textit{Medical Image Analysis (MedIA)}, \textit{The IEEE/CVF Conference on Computer Vision and Pattern Recognition (CVPR)}, \textit{International Conference on Medical Image Computing
and Computer Assisted Intervention (MICCAI)} and etc. This criterion guarantees that the selected literature has been subjected to rigorous peer review, ensuring its credibility.

\subsection{Literature Classification}
\begin{figure}[htbp]
    \centering
    \includegraphics[width=\textwidth]{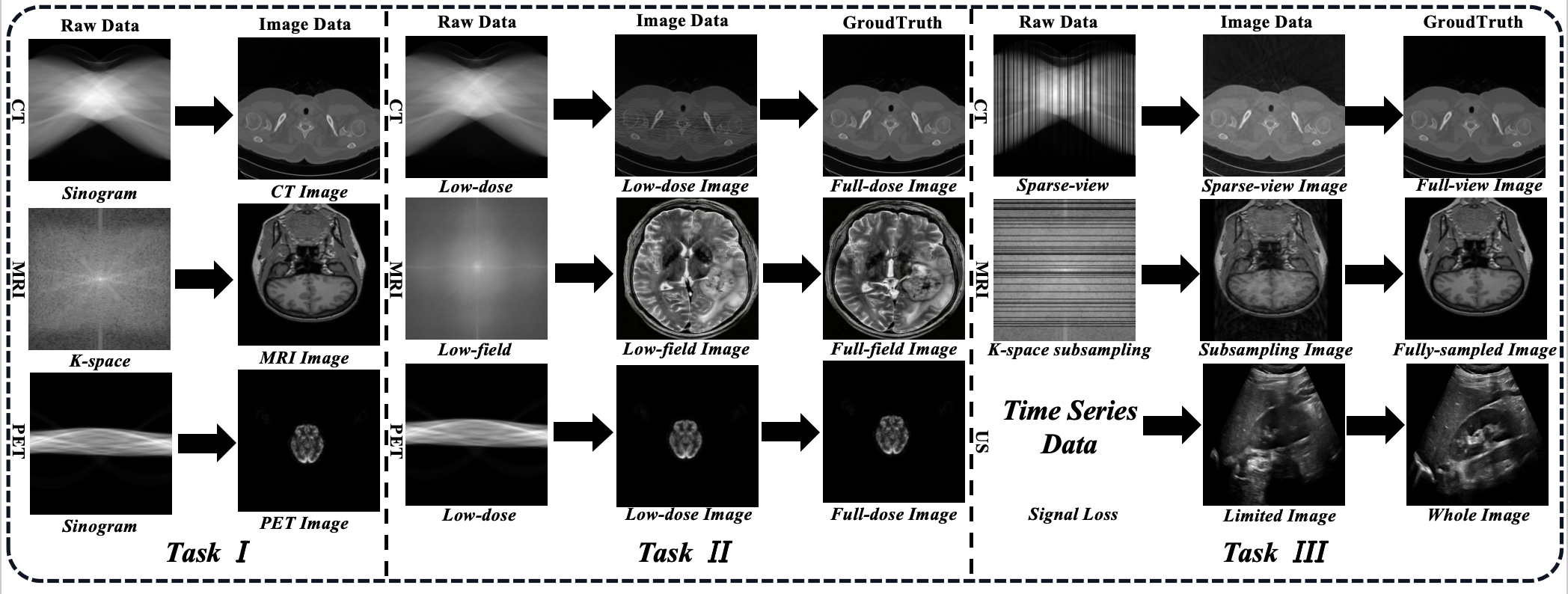} 
    \caption{Three distinct task types in radiological image reconstruction.}
    \label{fig:4}
\end{figure}

\begin{figure}[htbp]
    \centering
    \includegraphics[width=\linewidth]{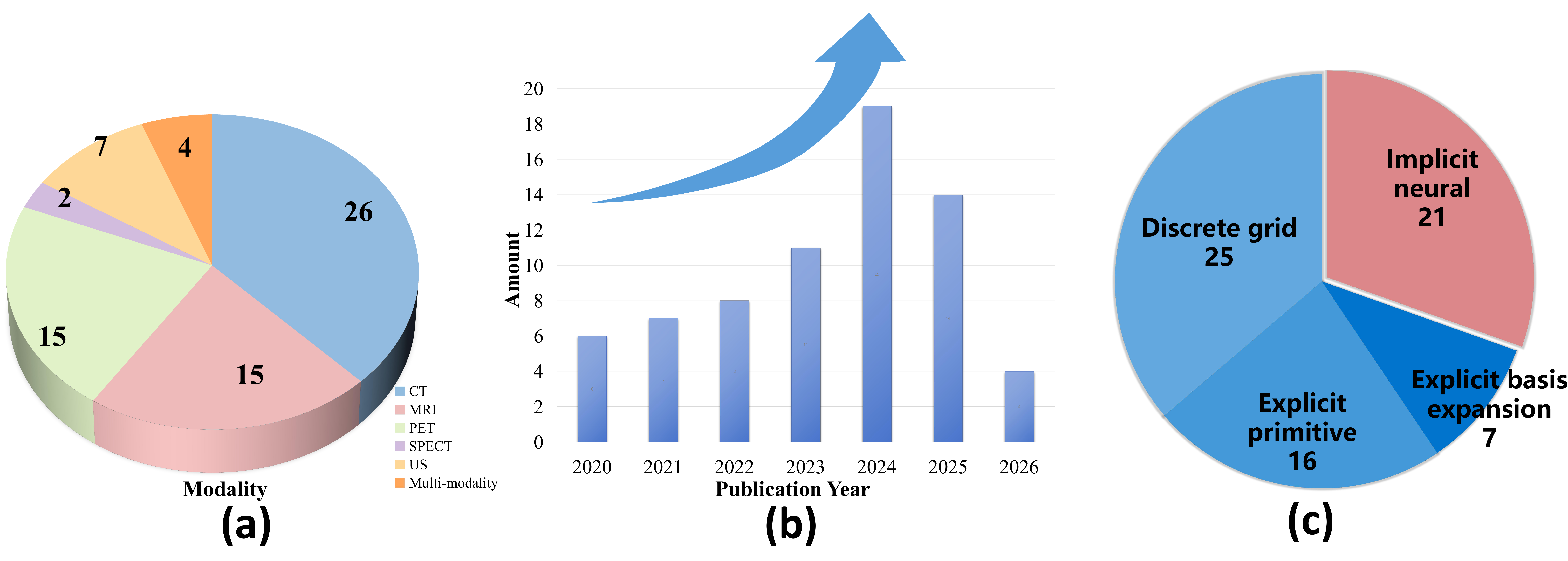} 
    \caption{Categorization results of the reviewed literature, including: \textbf{(a)} publication year, \textbf{(b)} imaging modalities, and \textbf{(c)} representation forms used in reconstruction methods.}
    \label{fig:3}
\end{figure}

There are multiple feasible criteria for classifying studies on 3D reconstruction in radiological imaging \citep{ahishakiye2021survey,sarmah2023survey}. One common approach is based on imaging modality, where studies are grouped according to CT, MRI, PET, SPECT, US, or multimodal imaging, as illustrated in Fig.~\ref{fig:3}\textbf{(b)}. Another approach focuses on the anatomical region of interest, such as brain, breast, cardiac, or whole-body imaging. A third and more task-oriented approach categorizes studies according to the reconstruction objective. In this review, we adopt three task categories, as shown in Fig.~\ref{fig:4}. \textbf{Task \textit{I}} refers to reconstructing 3D images directly from raw measurement data, such as projections, sinograms, k-space, list-mode data, or ultrasound RF/channel data, and is primarily aimed at recovering images from the acquisition domain while improving reconstruction efficiency. \textbf{Task \textit{II}} focuses on transforming low-dose or degraded images into normal-dose or higher-quality images, with the goal of improving image quality, suppressing noise, and reducing artifacts. \textbf{Task \textit{III}} refers to reconstructing images from incomplete or missing signals, aiming to recover complete anatomical information from sparse-view, limited-angle, undersampled, or novel-view observations. These three task categories are also used consistently in Table~\ref{tbl1} to organize the publicly available datasets reviewed in this paper.

While these classification schemes provide useful perspectives, they do not fully capture methodological differences from the viewpoint of artificial intelligence. From the perspective of 3D representation, reconstruction methods can be broadly divided into explicit and implicit representations. Explicit representations use directly parameterized forms, such as points, voxels, meshes, or Gaussian primitives, and are generally discrete in nature. In particular, Gaussian representations preserve a discrete primitive-based structure while maintaining local continuity within each primitive. Implicit representations, in contrast, model the reconstructed target through continuous functions, such as neural fields, and therefore represent imaging objects in a less direct but more flexible manner. From this AI-oriented perspective, we further re-examine 3D reconstruction methods in radiological imaging and classify the literature according to output representation, namely explicit and implicit reconstruction. The corresponding classification results are shown in Fig.~\ref{fig:3}\textbf{(c)}. \textbf{Accordingly, one of the key contributions of this work is that it classifies the literature from a representation-oriented perspective.}

\section{Datasets and metrics}
\label{sec3}
\subsection{Datasets}
In 3D reconstruction for radiological imaging, publicly available datasets play a crucial role \citep{li2023systematic}. High-quality and standardized datasets are essential for algorithm training, validation, and testing. Public benchmarks provide a common basis for fair comparison under consistent experimental settings, thereby facilitating method development and optimization \citep{ogier2022flamby}.

Compared with other medical imaging tasks, reconstruction datasets are often more difficult to acquire because they usually require either access to raw measurement data or paired acquisitions under comparable imaging conditions. In many cases, this means that the subject must be scanned in a highly consistent manner, with minimal variation in positioning, posture, acquisition protocol, and physiological state, so that degraded measurements and reference images can be meaningfully aligned. In addition, medical data collection is constrained by strict legal and regulatory requirements related to patient privacy \citep{kulynych2017clinical}, which further increases the cost and difficulty of dataset construction. Therefore, the availability of public datasets greatly lowers the barrier to reproducible research.

Table~\ref{tbl1} summarizes representative publicly available datasets for these task categories across CT, MRI, PET, SPECT, US, and multi-modality scenarios. And the \textit{Task} column is organized according to the three reconstruction task types illustrated in Fig.~\ref{fig:4}.

{\scriptsize
\setlength{\tabcolsep}{2pt}
\renewcommand{\arraystretch}{1.18}
\begin{longtable}{
>{\raggedright\arraybackslash}p{0.09\linewidth}
>{\raggedright\arraybackslash}p{0.09\linewidth}
>{\raggedright\arraybackslash}p{0.18\linewidth}
>{\raggedright\arraybackslash}p{0.18\linewidth}
>{\raggedright\arraybackslash}p{0.36\linewidth}
}
\caption{Representative public datasets for radiological image reconstruction and ultrasound image formation.}
\label{tbl1} \\
\toprule
\textbf{Modalities} &
\textbf{Task} &
\textbf{Dataset} &
\textbf{Size} &
\textbf{Detail} \\
\midrule

\multirow{6}{*}{CT}
& Task I / Task II
& AAPM Low Dose CT Grand Challenge \citep{mccollough2017low}
& 30 abdominal cases plus an ACR phantom
& A classic low-dose CT reconstruction benchmark providing quarter-dose and full-dose CT data for algorithm comparison under clinically relevant dose reduction. \\
\cmidrule(l){2-5}

& Task I / Task II
& LDCT-and-Projection-Data \citep{moen2021low}
& 299 clinical CT examinations
& A TCIA/Mayo dataset containing head, chest, and abdomen CT projection data in DICOM-CT-PD format, with routine-dose and simulated low-dose projections for CT reconstruction and denoising. \\
\cmidrule(l){2-5}

& Task I / Task II
& LoDoPaB-CT \citep{leuschner2021lodopab}
& 46,573 2D image--observation pairs
& A large low-dose CT benchmark derived from LIDC-IDRI, providing simulated low-photon-count CT observations and ground-truth images for inverse reconstruction. \\
\cmidrule(l){2-5}

& Task I / Task III
& SinoCT \citep{hooper2021impact}
& $>$9,000 head CT scans
& A large head CT dataset with reconstructed DICOM images and GE CatSim-simulated sinograms, suitable for sinogram-domain and reduced-projection CT reconstruction studies. \\
\cmidrule(l){2-5}

& Task II / Task III
& AAPM CT-MAR Challenge \citep{haneda2025aapm}
& 14,000 training cases, 1,000 test cases, and 29 final clinical scoring cases
& A CT metal artifact reduction benchmark providing paired metal-corrupted and metal-free images and sinograms, allowing image-domain, sinogram-domain, and hybrid reconstruction methods. \\
\cmidrule(l){2-5}

& Task I / Task III
& MORE \citep{wu2025more}
& 135 CT studies (65,575 slices)
& A multi-organ CT reconstruction benchmark covering 9 anatomies and 15 lesion types, designed for sparse-view CT reconstruction and cross-anatomy generalization evaluation. \\
\cmidrule(l){2-5}

& Task III
& Corona-Figueroa et al. \citep{corona2022mednerf}
& 25 CT volumes
& A paper-specific chest and knee CT-to-DRR corpus used for single- or few-view X-ray/CT projection reconstruction and 3D-aware radiological novel-view synthesis. \\
\midrule

\multirow{6}{*}{MRI}
& Task I / Task III
& fastMRI \citep{zbontar2018fastmri,knoll2020fastmri}
& 1,594 knee raw scans, 7,002 brain raw scans, plus prostate and breast subsets
& A large-scale MRI reconstruction benchmark releasing raw k-space and reference images for accelerated single-coil and multi-coil MRI reconstruction. \\
\cmidrule(l){2-5}

& Task I / Task III
& Calgary-Campinas Public Brain MR Raw k-space Dataset \citep{souza2018open,singh2023emerging}
& 167 3D T1-weighted brain MRI scans
& A public brain MRI raw k-space benchmark with 12-channel and 32-channel acquisitions for accelerated 3D MRI reconstruction. \\
\cmidrule(l){2-5}

& Task I / Task III
& OCMR \citep{chen2020ocmr}
& 165 fully sampled and 212 prospectively undersampled cardiac cine series
& An open-access cardiac MRI dataset providing multi-coil k-space data for fully sampled and prospectively undersampled cine reconstruction. \\
\cmidrule(l){2-5}

& Task I / Task III
& SKM-TEA \citep{desai2022skm}
& 155 qDESS knee MRI scans
& A quantitative knee MRI dataset with raw k-space, reconstructed DICOM images, tissue segmentations, and pathology labels for end-to-end accelerated MRI reconstruction evaluation. \\
\cmidrule(l){2-5}

& Task I / Task III
& CMRxRecon2024 \citep{wang2025cmrxrecon2024}
& 330 volunteers; 200/60/70 train/validation/test split
& A multi-contrast, multi-view, multi-coil cardiac MRI k-space benchmark covering cine, mapping, tagging, flow, and dark-blood acquisitions with multiple undersampling patterns. \\
\cmidrule(l){2-5}

& Task I / Task II
& M4Raw \citep{lyu2023m4raw}
& 183 healthy volunteers; $>$1,000 released volumes after quality control
& A low-field 0.3T multi-channel brain MRI raw k-space dataset with T1w, T2w, and FLAIR repetitions for denoising, averaging, and parallel-imaging reconstruction. \\
\midrule

\multirow{2}{*}{PET}
& Task II
& UDPET \citep{xue2025udpet}
& 1,447 whole-body $^{18}$F-FDG PET subjects
& A whole-body ultra-low-dose PET recovery benchmark providing low-statistics PET images at multiple dose-reduction factors aligned with full-dose PET images. \\
\cmidrule(l){2-5}

& Task I
& PETRIC \citep{da2026pet}
& Approximately 13 phantom datasets
& A PET rapid image reconstruction challenge dataset with phantom list-mode files, attenuation-correction files, normalization/calibration files, vendor reconstructions, and VOIs for PET reconstruction benchmarking. \\
\midrule

\multirow{3}{*}{US}
& Task I / Task III
& PICMUS \citep{liebgott2016plane}
& 4 core phantom datasets plus in-vivo carotid acquisitions
& A canonical plane-wave ultrasound image-formation benchmark providing RF/IQ data from 75 steered plane waves for coherent compounding and beamforming reconstruction. \\
\cmidrule(l){2-5}

& Task I / Task III
& CUBDL \citep{hyun2021deep}
& 576 ultrasound acquisition sequences
& A large open ultrasound channel-data benchmark containing simulated, phantom, and in-vivo acquisitions for deep-learning-based beamforming and image formation. \\
\cmidrule(l){2-5}

& Task III
& Wysocki et al. / Ultra-NeRF \citep{wysocki2024ultra}
& Synthetic liver sweeps and tracked spine-phantom sweeps
& A B-mode ultrasound novel-view and volumetric representation dataset for learning view-dependent US image synthesis from overlapping sweeps. \\
\midrule

\multirow{2}{*}{\makecell[l]{PET /\\MRI}}
& Task I
& Monash vis-fPET-fMRI \citep{jamadar2021task}
& 10 healthy adults
& An OpenNeuro simultaneous FDG-fPET/BOLD-fMRI dataset releasing unreconstructed PET list-mode source data, normalization and sinogram-related files, and reconstructed PET images for PET reconstruction pipeline development. \\
\cmidrule(l){2-5}

& Task I
& Monash DaCRA fPET-fMRI \citep{jamadar2022monash}
& 15 participants
& An OpenNeuro FDG-fPET/fMRI dataset comparing bolus, infusion, and bolus-infusion protocols and releasing unreconstructed PET list-mode data for offline dynamic PET reconstruction. \\
\midrule

\makecell[l]{SPECT /\\CT}
& Task I
& Lu-177 DOTATATE Projection Data \citep{brosch2024international,kurkowska2025international,brosch2023international,kurkowska2026international,uribe2021international}
& 2 patients scanned at 4 post-therapy time points
& A rare public human SPECT reconstruction dataset containing Lu-177 SPECT projection data and CT-based attenuation coefficient maps for quantitative SPECT reconstruction and dosimetry. \\

\bottomrule

\end{longtable}
}

\subsection{Metrics}

Evaluation metrics play a critical role in the study of medical image reconstruction by providing standardized assessment, guiding algorithm optimization, and validating clinical applicability. Scientifically sound and well-designed metrics enable an objective evaluation of reconstruction fidelity, detail preservation, and noise suppression, ensuring the practical utility of models across different application scenarios\citep{erickson2021magician}. Since different metrics focus on distinct aspects of image quality\citep{rainio2024evaluation}, a single metric is often insufficient to comprehensively reflect reconstruction performance. Therefore, a combination of multiple metrics is typically required for a more thorough and reliable evaluation. This review summarizes the commonly used metrics in 3D reconstruction for radiological image tasks, with the results presented in Table \ref{tbl2}.

\begin{table}[h]
\caption{Evaluation metrics for radiological image reconstruction}\label{tbl2}
\centering
\scriptsize 
\begin{tabularx}{\linewidth}{>{\raggedright\arraybackslash}p{0.22\linewidth}|>{\raggedright\arraybackslash}X|>{\raggedright\arraybackslash}X}  
\hline
\hline
Name & Formula & Function \\  
\hline
Mean Squared Error (MSE) $\downarrow$ & \( MSE = \frac{1}{N} \sum_{i=1}^{N} (x_i - y_i)^2 \) & Measures pixel-wise intensity differences.\\
\hline
Root Mean Squared Error (RMSE) $\downarrow$ & \( RMSE = \sqrt{MSE} \) & Square root of MSE. \\
\hline
Normalized Root Mean Square Error (NRMSE) $\downarrow$ & \( NRMSE = \frac{\sqrt{MSE}}{\max(x) - \min(x)} \) & Normalized MSE for different intensity ranges.  \\
\hline
Peak Signal-to-Noise Ratio (PSNR) $\uparrow$ &  \( PSNR = 10 \log_{10} \frac{\max(x)^2}{MSE} \) & Evaluates image quality based on signal strength.\\
\hline
Structural Similarity Index (SSIM) $\uparrow$ & \( SSIM(x,y) = \frac{(2\mu_x\mu_y + C_1)(2\sigma_{xy} + C_2)}{(\mu_x^2 + \mu_y^2 + C_1)(\sigma_x^2 + \sigma_y^2 + C_2)} \) & Measures structural similarity between images. Ranges from -1 to 1; \\
\hline
Learned Perceptual Image Patch Similarity (LPIPS) $\downarrow$ & $LPIPS(x, y) = \sum_l w_l \cdot ||F_l(x) - F_l(y)||_2^2$
& Measures perceptual similarity based on deep learning.\\
\hline
Mean Absolute Error (MAE) $\downarrow$ & \( MAE = \frac{1}{N} \sum_{i=1}^{N} |x_i - y_i| \) & Measures absolute intensity differences. \\
\hline
Feature Similarity Index (FSIM) $\uparrow$ & $FSIM = \frac{\sum_{x,y} PC_m(x,y) \cdot S_L(x,y) \cdot S_C(x,y)}{\sum_{x,y} PC_m(x,y)}$ & Evaluates perceptual quality with structural information.\\
\hline
Signal-to-Noise Ratio (SNR) $\uparrow$ & \( SNR = 10 \log_{10} \frac{\sum_{i=1}^{N} x_i^2}{\sum_{i=1}^{N} (x_i - y_i)^2} \) & Measures signal clarity relative to noise.  \\
\hline
Contrast-to-Noise Ratio (CNR) $\uparrow$ & \( CNR = \frac{|\mu_R - \mu_B|}{\sigma_B} \) & Evaluates contrast between regions of interest.  \\
\hline
Cosine Similarity (CS) $\uparrow$ & 
\( CS = \frac{\sum x_i y_i}{\sqrt{\sum x_i^2} \cdot \sqrt{\sum y_i^2}} \) & Measures the cosine of the angle between two image feature vectors. \\
\hline
Computation Time (Time) $\downarrow$ & Measured in seconds  & Evaluates computational efficiency.  \\
\hline
Rendering Speed (Speed) $\uparrow$ & Frames per second (FPS) & Measures the speed of rendering in frames per second. \\
\hline
Model Parameters (Para.) $\downarrow$ & Count of model parameters & Evaluates the complexity and size of the model. \\
\hline
\hline
\end{tabularx}
\normalsize
\end{table}

The definitions of the symbols used in Table \ref{tbl2} are as follows: \(x_i\) and \(y_i\) represent the pixel intensity values or feature vector components of the original and reconstructed images respectively, and \(N\) denotes the total number of pixels. The means and variances of images \(x\) and \(y\) are represented by \(\mu_x\), \(\mu_y\) and \(\sigma_x^2\), \(\sigma_y^2\) respectively, and \(\sigma_{xy}\) is their covariance. And \(\max(x)\) and \(\min(x)\) represent the maximum and minimum intensity values in the image.
The constants \( C_1 \) and \( C_2 \) are small positive stabilizing constants used in the SSIM calculation. In LPIPS, \( F_l(x) \) and \( F_l(y) \) represent the feature maps at layer \( l \), and \( w_l \) denotes the learned weight for each layer. FSIM is computed based on phase congruency \( PC_m(x,y) \), a contrast-based feature, along with luminance similarity \( S_L(x,y) \) and contrast similarity \( S_C(x,y) \).  For CNR, \( \mu_R \) and \( \mu_B \) represent the mean intensities of the regions of interest (ROI) and background. \( \sigma_B \) denotes the standard deviation of the background intensity.

Table \ref{tbl2} presents a comprehensive set of evaluation metrics for medical image reconstruction, categorized into pixel-wise error, signal and noise, perceptual similarity, and computational performance metrics. Pixel-wise error metrics, such as MSE, RMSE, NEMSE and MAE, quantify intensity differences between reconstructed and reference images. Signal and noise metrics, including PSNR, SNR and CNR, assess the clarity and contrast of images relative to noise levels. Perceptual and structural similarity metrics, like SSIM, LPIPS, FSIM and CS, capture the high-level structural and perceptual quality of images. Finally, computational performance metrics, such as Time, Speed and Para. evaluate the efficiency and feasibility of the reconstruction method in practical applications. Together, these indicators provide a balanced assessment of image quality, structural fidelity, and computational cost in medical image reconstruction.

\section{Discrete grid representations}
\label{sec4}

Among explicit representations, the most widely used form in radiological image reconstruction is the discrete grid representation. In this family, the target radiological image is directly stored on sampled spatial grids, and the reconstruction model predicts intensities on pixels, slices, or voxels. Let $\Omega \subset \mathbb{R}^3$ denote the imaging domain, and let $\lambda(\mathbf{r})$ denote the target signal at spatial location $\mathbf{r}\in\Omega$. A general grid representation can be written as
\begin{equation}
\lambda(\mathbf{r}) \approx \sum_{m=1}^{M} x_m \phi_m(\mathbf{r}),
\end{equation}
where $\phi_m(\mathbf{r})$ are grid-aligned indicator basis functions and $x_m$ are explicitly stored coefficients. In this sense, both slice-based and volume-based methods are explicit grid representations, differing mainly in whether the grid is processed plane-wise or jointly in 3D.

Compared with implicit representations, grid representations are intuitive, easy to implement, and naturally compatible with existing imaging pipelines. For this reason, they were initially the dominant form in AI-assisted radiological image reconstruction. This representation is widely applied in CT, MRI, PET and SPECT imaging, with potential for multi-modality applications. More details are shown in Table \ref{tbl4}.
{\scriptsize
\setlength{\tabcolsep}{3pt}
\renewcommand{\arraystretch}{1.05}
\begin{longtable}{
>{\raggedright\arraybackslash}p{0.13\linewidth}
>{\raggedright\arraybackslash}p{0.10\linewidth}
>{\raggedright\arraybackslash}p{0.24\linewidth}
>{\centering\arraybackslash}p{0.08\linewidth}
>{\raggedright\arraybackslash}p{0.35\linewidth}
}
\caption{Representative methods under the \textbf{Discrete grid representations} family. Slice-wise, Multi-plane, and Volume correspond to the representations introduced in Section~\ref{sec4}.}
\label{tbl4} \\
\toprule
Rep. & Mod. & Paper & Year & Detail \\
\midrule

\multirow{9}{*}{Slice-wise}

& \multirow{3}{*}{CT}
& CTTR \citep{shi2022dual}
& 2022
& A dual-domain transformer for sparse-view CT reconstruction. It restores sinogram and image-domain information while the final reconstruction remains an explicit CT grid. \\
\cmidrule(l){3-5}

&
& DuDoTrans \citep{wang2022dudotrans}
& 2022
& A dual-domain transformer that restores sparse-view sinograms and reconstructs CT images by combining projection-domain and image-domain information. \\
\cmidrule(l){3-5}

&
& CoreDiff \citep{gao2023corediff}
& 2023
& A diffusion-based low-dose CT reconstruction method that uses a mean-preserving degradation process and image-domain restoration to suppress noise and preserve texture. \\
\cmidrule(l){2-5}
& \multirow{4}{*}{MRI}
& Du et al. \citep{du2020super}
& 2020
& A residual CNN-based super-resolution method for anisotropic 3D MRI. Although the task targets 3D MRI, the network processes grid samples and outputs rasterized image slices. \\
\cmidrule(l){3-5}

&
& VarNet \citep{sriram2020end}
& 2020
& An end-to-end variational network for accelerated MRI reconstruction. It became a strong fastMRI baseline and reconstructs images on explicit sampled grids. \\
\cmidrule(l){3-5}

&
& PCNN \citep{shen2021rapid}
& 2021
& A perceptual complex neural network for undersampled non-Cartesian cine MRI reconstruction, using complex convolutions and perceptual loss on discrete image grids. \\
\cmidrule(l){3-5}

&
& SLATER \citep{korkmaz2022unsupervised,9695412}
& 2022
& An unsupervised adversarial transformer-based MRI reconstruction method that learns an image prior and performs zero-shot reconstruction on sampled image grids. \\

\cmidrule(l){2-5}
& CT/MRI
& MambaMIR \citep{huang2025enhancing}
& 2025
& A Mamba-based medical reconstruction framework with arbitrary masking and uncertainty estimation, applied to fast MRI and sparse-view CT reconstruction. \\

\cmidrule(l){2-5}
& \makecell[l]{CT/MRI\\ /PET}
& Restore-RWKV \citep{yang2024restore}
& 2024
& A medical image restoration model based on recurrent WKV attention and omnidirectional token shift. It handles CT, MRI, and PET restoration on discrete image grids. \\

\midrule

\multirow{2}{*}{Multi-plane}
& CT
& Corona-Figueroa et al. \citep{corona2024repeat}
& 2024
& A method that repeats and concatenates 2D X-ray views into a 3D volume representation and uses neural optimal transport to retain cross-view information. \\
\cmidrule(l){2-5}
& MRI
& MADGAN \citep{han2021madgan}
& 2021
& A GAN-driven method that uses multiple adjacent brain MRI slices to improve reconstruction and anomaly detection, making it more naturally a slice-stack grid representation. \\

\midrule
\multirow{16}{*}{Volume}
& \multirow{6}{*}{CT}
& HDNet \citep{hu2020hybrid}
& 2020
& A hybrid-domain neural network for sparse-view cone-beam CT reconstruction, combining projection and image-domain information in a volumetric grid output. \\
\cmidrule(l){3-5}

&
& C$^{2}$RV \citep{Lin_2024_CVPR}
& 2024
& A cross-regional and cross-view learning framework for sparse-view CBCT reconstruction, using multi-scale volumetric representation and cross-view attention. \\
\cmidrule(l){3-5}

&
& Geometry-aware attenuation learning \citep{liu2024geometry}
& 2024
& A sparse-view CBCT reconstruction method that encodes multi-view projections, back-projects features into 3D space, and decodes a voxel volume. \\
\cmidrule(l){3-5}

&
& DCT-Net \citep{zhang2025dct}
& 2025
& A dual-branch CT reconstruction network from orthogonal X-rays, combining diffusion-based augmentation and perceptual contrastive learning for volumetric CT recovery. \\
\cmidrule(l){2-5}

& \multirow{2}{*}{MRI}
& CINENet \citep{kustner2020cinenet}
& 2020
& A 4D deep reconstruction framework for isotropic 3D cine MRI from single breath-hold acquisitions, reconstructing volumetric image sequences. \\
\cmidrule(l){3-5}

&
& Shaul et al. \citep{shaul2020subsampled}
& 2020
& A GAN-based undersampled MRI reconstruction framework combining U-Net and adversarial learning to recover high-quality grid-based MRI volumes. \\
\cmidrule(l){2-5}

& \multirow{6}{*}{PET}
& DirectPET \citep{whiteley2020directpet}
& 2020
& An end-to-end PET reconstruction method with a Radon inversion layer, designed for fast reconstruction from projection data to image grids. \\
\cmidrule(l){3-5}

&
& Xie et al. \citep{xie2021anatomically}
& 2021
& A supervised co-learning CNN framework that extracts PET and CT features and integrates them into constrained PET reconstruction to improve lesion contrast. \\
\cmidrule(l){3-5}

&
& MEaTransGAN \citep{wang20243d}
& 2024
& A 3D multi-modality Transformer-GAN that reconstructs standard-dose PET from low-dose PET and T1-weighted MRI. \\
\cmidrule(l){3-5}

&
& Singh et al. \citep{melba:2024:001:singh}
& 2024
& A fully 3D PET reconstruction framework using score-based generative models and PET-specific sampling to improve robustness and image quality. \\
\cmidrule(l){3-5}
& 
& Cycle-DCN \citep{hou2025cycle}
& 2025
& A cycle-constrained adversarial denoising CNN for 3D low-dose PET image denoising, validated on large datasets with reader study and real low-dose data. The method operates on discrete volumetric PET images rather than continuous or primitive-based representations. \\
\cmidrule(l){3-5}
&
& DDPET-3D \citep{xie2026doseaware}
& 2026
& A dose-aware diffusion model for 3D low-dose PET denoising that generates consistent volumetric PET images across varying dose levels and improves image quality in ultra-low-dose settings. \\
\cmidrule(l){2-5}
& SPECT
& SPECTnet \citep{shao2021spectnet}
& 2021
& A deep learning method for direct SPECT reconstruction from projection data to high-resolution activity images. \\
\cmidrule(l){2-5}
& CT/MRI
& DiffusionMBIR \citep{chung2023solving}
& 2023
& A method combining pretrained 2D diffusion priors with model-based iterative reconstruction for sparse-view CT, limited-angle CT, and compressed sensing MRI. \\
\bottomrule
\end{longtable}
}

\subsection{Slice-wise raster representation}
\label{sec:slice}

The slice-wise raster representation reconstructs a 3D image as an ordered stack of 2D slices:
\begin{equation}
\mathbf{X}=\{X^{(k)}\}_{k=1}^{D}, \qquad X^{(k)}\in\mathbb{R}^{H\times W},
\end{equation}
which is equivalent to a 3D tensor $\mathbf{X}\in\mathbb{R}^{H\times W\times D}$ but is processed one slice at a time. 

The slice-based method reconstructs images in a single direction, with early deep learning approaches relying on CNN-based U-Net \citep{ronneberger2015u} architectures. Du et al. \citep{du2020super} introduced a CNN model with residual learning and skip connections, effectively addressing gradient vanishing and achieving superior PSNR and SSIM compared to traditional methods. Their model also enabled multi-modality super-resolution by reconstructing T2-weighted MRI from T1-weighted MRI training. With the rise of ViTs in computer vision \citep{dosovitskiy2020image}, Shi et al. proposed CTTR \citep{shi2022dual}, a Transformer-based sparse-view CT reconstruction method that reduces artifacts and detail loss. CTTR outperformed CNN-based methods in PSNR and SSIM under extremely sparse-view conditions.

Mamba \citep{gu2023mamba}, a state-space model for efficient sequence processing, has recently gained attention. Building on this, MambaMIR \citep{huang2025enhancing} adapts Mamba for medical image reconstruction, leveraging its linear complexity, global receptive fields, and dynamic weighting. It introduces an Arbitrary-Masked Mechanism, enhancing its suitability for medical imaging, and incorporates Monte Carlo-based uncertainty estimation. Experiments show that MambaMIR achieves SOTA or superior performance in Fast MRI and Sparse-View CT tasks, effectively handling long-sequence data while optimizing computational efficiency for large-scale medical image reconstruction.

Although Mamba has unidirectionality issues, making it difficult to model both global and local dependencies in 2D images efficiently, newer sequence models provide alternative solutions. The Receptance Weighted Key Value (RWKV) model \citep{peng2023rwkv} introduced WKV attention, enabling long-range dependencies while maintaining linear computational complexity. Restore-RWKV \citep{yang2024restore} is the first RWKV-based medical image restoration model, and incorporates a Token Shift layer to enhance local feature capture. It is designed to address high computational complexity and insufficient global information capture in high-resolution medical imaging.

The main advantage of slice-wise raster reconstruction lies in its versatility and low memory burden, as the model can be adapted to different imaging modalities with relatively simple implementation. However, such methods overlook longitudinal continuity, often leading to inter-slice inconsistency, staircase artifacts, or structural discontinuity across the reconstructed 3D volume.

\subsection{Multi-plane and slice-stack raster representation}
\label{sec:multiplane}

A practical extension of the slice-wise paradigm is to reconstruct multiple adjacent slices jointly or to fuse information from multiple anatomical planes, such as axial, coronal, and sagittal views. In this case, the target is still represented explicitly on sampled planes, but the effective local support becomes larger:
\begin{equation}
\mathbf{X}_{\mathrm{local}}=\{X^{(k-s)},\ldots,X^{(k)},\ldots,X^{(k+s)}\}.
\end{equation}
This form remains a grid representation because the reconstructed signal is still stored on discrete sampled planes rather than being defined as a continuous coordinate function.

Compared with purely slice-wise processing, multi-plane and slice-stack representations partially recover 3D spatial continuity while retaining lower computational cost than full 3D reconstruction. Conceptually, they should not be regarded as an independent ontological representation category, but as an intermediate granularity within explicit grid reconstruction.

Representative examples have appeared in both slice-stack and cross-view fusion settings. For instance, MADGAN jointly exploits multiple adjacent brain MRI slices to improve reconstruction quality and anomaly detection, making it more naturally a slice-stack grid representation rather than a purely slice-wise raster formulation \cite{han2021madgan}. Although these methods incorporate richer inter-slice or inter-view context than conventional 2D reconstruction models, their outputs are still defined on discrete sampled planes or stacked grids, and therefore remain within the family of explicit grid representations.

\subsection{Volumetric grid representation}
\label{sec:volume}

Later, volume-based reconstruction methods emerged. Unlike slice-based approaches, volumetric grid representations jointly reconstruct the entire 3D image:
\begin{equation}
\mathbf{X}\in\mathbb{R}^{H\times W\times D},
\end{equation}
or equivalently,
\begin{equation}
\lambda(\mathbf{r}) \approx \sum_{i=1}^{H}\sum_{j=1}^{W}\sum_{k=1}^{D} x_{ijk}\,\phi^{\mathrm{vox}}_{ijk}(\mathbf{r}),
\end{equation}
where $\phi^{\mathrm{vox}}_{ijk}(\mathbf{r})$ denotes the indicator basis of voxel $(i,j,k)$. This representation is applicable to all three-dimensional imaging modalities, including CT, MRI, PET, SPECT and multi-modality. More details are shown in Table \ref{tbl4}.

Volume-based representations describe 3D structures using numerous small volumetric elements \citep{hohne1992volume,chen2012volume}. This approach enables high-fidelity, detail-rich radiology image reconstruction through direct processing of volumetric data. In medical applications, this approach precisely captures complex anatomical structures within the human body.

Although volume-based reconstruction has employed UNet networks for generation, this section focuses on more advanced generative models. GANs train through an adversarial process between a generator and a discriminator \citep{goodfellow2014generative}, allowing the generator to produce high-quality data that matches the real data distribution.

In radiology image reconstruction, Shaul et al. \citep{shaul2020subsampled} proposed a GAN-based undersampled MRI reconstruction method that accelerates MRI acquisition while maintaining anatomical image quality. They optimized reconstruction with fidelity loss, image-quality loss, and adversarial loss, generating more realistic and natural MRI images. Experiments demonstrated that this method can reconstruct high-quality brain MRI with fivefold acceleration.

Similarly, X2CT-GAN \citep{ying2019x2ct} leverages GANs to reconstruct 3D CT images from biplanar X-rays, reducing radiation exposure and the high costs of traditional CT scans. By incorporating a feature fusion module to integrate information from different views, the method enhances reconstruction quality. Experiments showed a 4dB improvement in PSNR and superior SSIM performance compared to CNN-based methods.

Despite GANs' effectiveness, they suffer from mode collapse and unstable training \citep{kushwaha2020study}. Diffusion models \citep{sohl2015deep,ho2020denoising,song2020denoising,rombach2022high}, optimized via Markov chains and variational inference, stabilize the generation process by progressively adding noise and learning the denoising process, achieving broader coverage of the real data distribution.

Chung et al. introduced DiffusionMBIR \citep{chung2023solving}, which combines pretrained 2D diffusion models with model-based iterative reconstruction for sparse-view CT, limited-angle CT, and compressed sensing MRI. By integrating a 2D diffusion prior with 3D total variation regularization, the approach significantly improves 3D reconstruction quality. Experimental results show that DiffusionMBIR maintains high-quality reconstruction even with extreme data sparsity, such as two-view CT reconstruction, outperforming existing diffusion and deep learning methods in PSNR and SSIM.

In the PET domain, Gong et al. \citep{gong2024pet} proposed a denoising diffusion probabilistic model (DDPM)-based method for PET image denoising. By progressively adding noise and iteratively denoising during the reverse process, this approach achieves high-quality PET image reconstruction. Results indicate that DDPM outperforms traditional methods such as non-local means (NLM), UNet, and GANs in both PSNR and SSIM.

Due to its ability to explicitly represent 3D information with high accuracy \citep{stull2014accuracy,zheng2010effective}, the volumetric grid representation is widely adopted. However, its computational complexity poses challenges for real-time imaging \citep{xu2007real,engel2004real}. Future research will continue striving for fast and accurate volumetric grid reconstruction.

\section{Explicit basis expansion representations}
\label{sec5}

Beyond grid representation, many radiological image reconstruction methods can be described more generally through basis expansion. In this formulation, the target image is represented by a finite set of coefficients over predefined basis functions:
\begin{equation}
\lambda(\mathbf{r}) = \sum_{m=1}^{M} \alpha_m \varphi_m(\mathbf{r}),
\label{eq:basis_general}
\end{equation}
where $\varphi_m(\mathbf{r})$ denotes the $m$-th basis function and $\alpha_m$ is its corresponding coefficient. As long as the reconstruction target is parameterized by a finite-dimensional coefficient vector, the representation remains explicit.

Under this view, grid-based reconstruction can be regarded as a special case of basis expansion, in which the basis functions are voxel indicators on a regular sampling lattice. However, explicit basis expansion representations are broader than regular grids, because they also include smoother local bases and transformed coefficient models. Their common feature is that the reconstructed image is still encoded by a finite set of coefficients rather than by a continuous coordinate-query function. This representation was widely applied in PET imaging. More details are shown in Table \ref{tbl5}.

{\scriptsize
\setlength{\tabcolsep}{3pt}
\renewcommand{\arraystretch}{1.05}
\begin{longtable}{
>{\raggedright\arraybackslash}p{0.14\linewidth}
>{\raggedright\arraybackslash}p{0.11\linewidth}
>{\raggedright\arraybackslash}p{0.27\linewidth}
>{\centering\arraybackslash}p{0.08\linewidth}
>{\raggedright\arraybackslash}p{0.30\linewidth}
}
\caption{Representative methods under the \textbf{Explicit basis expansion representations} family. Local basis and Kernel coeff. correspond to the representations introduced in Section~\ref{sec5}.}
\label{tbl5} \\
\toprule
Rep. & Mod. & Paper & Year & Detail \\
\midrule

\multirow{1}{*}{Local basis}
& PET
& Mandeville et al. \citep{mandeville2024partial}
& 2024
& A recent PET study using B-spline basis functions for voxel-wise partial volume correction, showing that local basis representations remain relevant in reconstruction-related medical imaging problems. \\
\midrule

\multirow{6}{*}{Kernel coeff.}
& \multirow{5}{*}{PET}
& Gong et al. \citep{gong2021direct}
& 2021
& A direct parametric PET reconstruction method combining nonlocal deep image prior, a kernel matrix layer, and kinetic-model convolution. The reconstructed object remains a finite coefficient-based image. \\
\cmidrule(l){3-5}

&
& Li et al. \citep{li2022deep}
& 2022
& A PET reconstruction framework that connects kernel representation with learnable deep mappings, improving dynamic PET reconstruction over empirical kernel construction. \\
\cmidrule(l){3-5}

&
& Neural KEM \citep{li2022neural}
& 2023
& A kernel expectation-maximization method with a deep coefficient prior, designed to improve low-count PET reconstruction while retaining explicit coefficient-space representation. \\
\cmidrule(l){3-5}

&
& Guo et al. \citep{guo2023pet}
& 2023
& A PET reconstruction method that jointly regularizes image space and kernel space to reduce mismatch between anatomical priors and PET tracer distributions. \\
\cmidrule(l){3-5}

&
& Deidda et al. \citep{deidda2023triple}
& 2023
& A multimodal kernel-based reconstruction framework incorporating SPECT, PET, and CT side information, showing that kernel coefficient ideas extend beyond PET-only reconstruction. \\
\cmidrule(l){2-5}

& SPECT
& Deidda et al. \citep{deidda2022hybrid}
& 2022
& A hybrid kernelised expectation-maximisation method for Bremsstrahlung SPECT reconstruction, improving image resolution and noise properties with CT guidance. \\

\bottomrule

\end{longtable}
}

\subsection{Local basis representations}
\label{sec:local_basis}

A natural extension of grid representation is to replace voxel indicators with smoother local basis functions:
\begin{equation}
\lambda(\mathbf{r}) = \sum_{m=1}^{M} \alpha_m \varphi_m(\mathbf{r}),
\end{equation}
where each $\varphi_m(\mathbf{r})$ has local spatial support but is not restricted to a hard voxel indicator. Compared with regular grid elements, such bases provide smoother spatial transitions and a more compact description of local image structure.

A representative example is the Kaiser-Bessel basis, commonly referred to as the blob basis \citep{lewitt1990multidimensional,lewitt1992alternatives,defrise2005image}. Blob functions are radially symmetric local bases with finite support and have long been studied as an alternative to conventional voxel indicators in tomographic reconstruction. Compared with voxel elements, they better model spatial continuity and often improve the trade-off between resolution and noise, although the overlap between neighboring bases makes projection and backprojection more computationally demanding \citep{defrise2005image}.

Early work by Lewitt established generalized Kaiser-Bessel window functions and explicitly highlighted alternatives to voxels in iterative reconstruction. Later PET studies further demonstrated their practical value. For example, Cabello et al. used spherical basis functions on a polar grid for 3D PET reconstruction \citep{cabello2012high}, while Leemans et al. evaluated blob-based time-of-flight PET reconstruction in hybrid brain PET/MR imaging \citep{leemans2015qualitative}. Related local-basis formulations have also been explored in CT, including optimized Kaiser-Bessel window functions for tomographic discretization \citep{nilchian2015optimized}. Therefore, local basis representations can be regarded as a general explicit strategy for tomographic reconstruction, although their practical adoption has been most visible in PET and related emission imaging systems. From a representation perspective, the unknown remains a finite coefficient vector, so the model is still explicit.

\subsection{Kernel coefficient representations}
\label{sec:kernel_basis}

Another important form of explicit basis expansion is to reconstruct the image through an intermediate coefficient space:
\begin{equation}
\mathbf{x} = K \boldsymbol{\alpha},
\label{eq:kernel_coeff}
\end{equation}
where $\mathbf{x}$ is the reconstructed image, $K$ is a predefined or learned kernel matrix, and $\boldsymbol{\alpha}$ is the coefficient vector to be estimated. In this formulation, the reconstruction target is generated from a finite set of coefficients through a linear mapping, rather than being predicted directly in image space.

This representation remains explicit because the unknown is still finite-dimensional. A representative milestone is the PET kernel method proposed by Wang and Qi, which models the PET image through kernel-induced coefficient expansion instead of direct voxel-wise estimation \citep{wang2014pet}. This idea was later extended to anatomically aided PET reconstruction, where MRI-derived structural information is used to construct kernel features for PET recovery \citep{hutchcroft2016anatomically}.

Kernel coefficient representations have also been applied to dynamic and parametric PET. For example, direct Patlak reconstruction with MRI-informed kernels estimates parametric images directly from dynamic PET data \citep{gong2017direct}, while spatiotemporal kernel reconstruction further incorporates temporal correlations to improve high-temporal-resolution dynamic PET reconstruction \citep{wang2018high}. More recent studies combine kernel representations with neural networks, including nonlocal deep image prior, deep kernel representation, Neural KEM, and kernel-space composite regularization \citep{gong2021direct,li2022neural,li2022deep,guo2023pet}. Although these methods may involve learned modules, the reconstructed target is still encoded by a finite coefficient vector or coefficient image, and therefore remains an explicit basis expansion representation rather than an implicit neural field.

\section{Explicit primitive representations}
\label{sec6}

Different from grid representations and basis expansion representations, primitive representations reconstruct the target image by a finite set of explicitly parameterized local elements. In this formulation, the image is not described on a uniform sampling lattice, but is approximated by adaptive primitives with learnable parameters:
\begin{equation}
\lambda(\mathbf{r}) \approx \sum_{k=1}^{K} \psi_k(\mathbf{r};\Theta_k),
\label{eq:primitive_general}
\end{equation}
where $\psi_k$ denotes the $k$-th primitive and $\Theta_k$ denotes its explicit parameters. Since all primitive parameters are directly stored and optimized, this representation remains explicit.

Compared with regular grids, primitive representations provide adaptive spatial support and can be more efficient in rendering and optimization. In radiological image reconstruction, the most representative primitive forms include Gaussian primitives and simplex mesh primitives. Among these representations, Gaussian-based methods are widely applied in CT and expanding to MRI and US, showing multimodal potential. In contrast, simplex mesh-based methods have received less attention. See Table \ref{tbl6} for details.

{\scriptsize
\setlength{\tabcolsep}{3pt}
\renewcommand{\arraystretch}{1.05}
\begin{longtable}{
>{\raggedright\arraybackslash}p{0.14\linewidth}
>{\raggedright\arraybackslash}p{0.10\linewidth}
>{\raggedright\arraybackslash}p{0.25\linewidth}
>{\centering\arraybackslash}p{0.08\linewidth}
>{\raggedright\arraybackslash}p{0.33\linewidth}
}
\caption{Representative methods under the \textbf{Explicit primitive representations} family. Gaussian and Simplex mesh correspond to the representations introduced in Section~\ref{sec6}.}
\label{tbl6} \\
\toprule
Rep. & Mod. & Paper & Year & Detail \\
\midrule

\multirow{16}{*}{Gaussian}
& \multirow{8}{*}{CT}
& DIF-Gaussian \citep{lin2024learning}
& 2024
& A Gaussian-based sparse-view CBCT reconstruction method that learns 3D Gaussian features and optimizes them for extremely sparse projection settings. \\
\cmidrule(l){3-5}

&
& R$^{2}$-Gaussian \citep{zha2024r}
& 2024
& A radiative Gaussian splatting framework for tomographic reconstruction, designed to reduce integration bias and improve sparse-view reconstruction quality. \\
\cmidrule(l){3-5}

&
& GaSpCT \citep{nikolakakis2024gaspct}
& 2024
& A Gaussian splatting framework for CT projection view synthesis from limited projections, improving efficiency compared with dense voxel and NeRF-style representations. \\
\cmidrule(l){3-5}

&
& DDGS-CT \citep{gao2024ddgsct}
& 2024
& A direction-disentangled Gaussian splatting method for realistic CT volume rendering, decomposing radiosity into isotropic and directional components. \\
\cmidrule(l){3-5}

&
& X-Gaussian \citep{cai2025radiative}
& 2025
& A radiative Gaussian splatting method for efficient X-ray novel-view synthesis, eliminating view dependency and improving training and inference speed. \\
\cmidrule(l){3-5}

&
& X-GRM \citep{liu2025x}
& 2025
& A large Gaussian reconstruction model that decodes sparse-view X-rays into a voxel-based Gaussian representation for feed-forward CT reconstruction. \\
\cmidrule(l){3-5}

&
& 3DGR-CT \citep{li20253dgr}
& 2025
& A sparse-view CT reconstruction method that initializes Gaussian primitives from FBP and updates them adaptively during self-supervised reconstruction. \\
\cmidrule(l){3-5}

&
& Fu et al. \citep{fu2025dynamic}
& 2025
& A 4D-CBCT reconstruction method that combines Gaussian primitives with deformation modeling to represent respiratory motion and dynamic anatomy. \\
\cmidrule(l){2-5}

& \multirow{3}{*}{MRI}
& CauGD2-Net \citep{hou2025causality}
& 2025
& A causality-driven dual-domain MRI reconstruction network enhanced by Gaussian-splatting-inspired spatial modeling to reduce artifacts and improve consistency. \\
\cmidrule(l){3-5}

&
& 3DGSMR \citep{peng2025three}
& 2025
& A 3D MRI reconstruction framework from undersampled k-space that represents the target MR volume with explicit 3D Gaussian primitives. \\
\cmidrule(l){3-5}

&
& M-Gaussian \citep{zheng2026m}
& 2026
& A magnetic Gaussian framework for slice-to-volume MRI reconstruction that adapts 3D Gaussian primitives to efficient multi-stack MRI reconstruction. \\
\cmidrule(l){2-5}

& PET
& GR-Diffusion \citep{geng2026gr}
& 2026
& A whole-body PET reconstruction framework that combines explicit 3D Gaussian representation with diffusion-based refinement to improve global consistency and local detail recovery. \\
\cmidrule(l){2-5}

& \multirow{3}{*}{US}
& UltraGauss \citep{eid2025ultragauss}
& 2025
& An ultrasound-specific Gaussian splatting framework for fast 3D ultrasound volume reconstruction, modeling probe-plane intersections to better match acoustic image formation. \\
\cmidrule(l){3-5}

&
& UltraGS \citep{yang2025ultrags}
& 2025
& A depth-aware Gaussian splatting framework for ultrasound novel view synthesis, incorporating ultrasound-specific rendering effects such as attenuation, reflection, and scattering. \\
\cmidrule(l){3-5}

&
& UltraG-Ray \citep{duelmer2026ultrag}
& 2026
& A physics-based ultrasound rendering method built on a learnable 3D Gaussian field, designed for realistic novel-view B-mode synthesis. \\
\midrule

\multirow{1}{*}{Simplex mesh}

& PET
& Lesonen et al. \citep{lesonen2024anatomy}
& 2024
& A PET reconstruction method using anatomy-guided multi-resolution triangular meshes to achieve a favorable balance between accuracy and computational cost. \\

\bottomrule

\end{longtable}
}

\subsection{Gaussian primitive representation}
\label{sec:gr}

Gaussian primitive representation has recently become an increasingly important explicit representation family for radiological image reconstruction. Inspired by 3D Gaussian splatting (3DGS) \citep{kerbl20233d}, these methods represent the target anatomy or imaging volume using a finite set of learnable Gaussian primitives rather than a dense voxel grid or a purely implicit coordinate network. As summarized in Table~\ref{tbl6}, recent Gaussian-based reconstruction methods have expanded from sparse-view CT and CBCT reconstruction to MRI reconstruction, whole-body PET reconstruction, and ultrasound volume reconstruction or novel-view synthesis \citep{lin2024learning,zha2024r,cai2025radiative,hou2025causality,peng2025three,geng2026gr,eid2025ultragauss}.

Mathematically, a Gaussian primitive representation can be written as
\begin{equation}
\lambda(\mathbf{r}) \approx \sum_{k=1}^{K} w_k\,\mathcal{G}(\mathbf{r};\boldsymbol{\mu}_k,\Sigma_k),
\label{eq:gaussian_primitive}
\end{equation}
where $w_k$ denotes the primitive weight or density coefficient, $\boldsymbol{\mu}_k$ denotes the center, and $\Sigma_k$ denotes the covariance matrix of the $k$-th Gaussian primitive. Compared with regular voxel grids, Gaussian primitives provide adaptive and anisotropic local support. They can concentrate representational capacity around informative structures while avoiding uniform discretization of the entire reconstruction domain. Since the reconstructed object is explicitly stored as a finite set of optimized Gaussian elements, this family remains an explicit primitive representation even when differentiable rendering or neural optimization is used.

In CT reconstruction, Gaussian representations have been explored mainly for sparse-view and projection-based reconstruction. DIF-Gaussian learns 3D Gaussian features for sparse-view CBCT reconstruction \citep{lin2024learning}, while R$^{2}$-Gaussian introduces radiative Gaussian splatting to reduce integration bias in tomographic reconstruction \citep{zha2024r}. Other methods further improve the representation and rendering process. For example, GaSpCT uses Gaussian splatting for CT projection view synthesis from limited projections \citep{nikolakakis2024gaspct}, DDGS-CT decomposes radiosity into isotropic and directional components for more realistic CT volume rendering \citep{gao2024ddgsct}, and X-Gaussian removes view dependency for efficient X-ray novel-view synthesis \citep{cai2025radiative}. More recent works such as X-GRM, 3DGR-CT, and dynamic Gaussian representations further investigate feed-forward reconstruction, FBP-initialized self-supervised optimization, and 4D-CBCT motion modeling \citep{liu2025x,li20253dgr,fu2025dynamic}.

Beyond CT, Gaussian primitive representations have also been adopted in other imaging modalities. In MRI, CauGD2-Net introduces Gaussian-splatting-inspired spatial modeling into a causality-driven dual-domain reconstruction network \citep{hou2025causality}, 3DGSMR represents undersampled 3D MRI volumes with explicit Gaussian primitives \citep{peng2025three}, and M-Gaussian adapts magnetic 3D Gaussians to slice-to-volume MRI reconstruction \citep{zheng2026m}. In PET, GR-Diffusion combines explicit 3D Gaussian representation with diffusion refinement for whole-body PET reconstruction \citep{geng2026gr}. In ultrasound, UltraGauss applies ultrasound-specific Gaussian splatting to fast 3D ultrasound volume reconstruction, while UltraGS and UltraG-Ray extend Gaussian-based representations to ultrasound novel-view synthesis and physics-based rendering \citep{eid2025ultragauss,yang2025ultrags,duelmer2026ultrag}.

Although many Gaussian-based methods borrow ideas from neural radiance fields or differentiable rendering, such as ray integration, view synthesis, and rendering-based supervision, their representation type is still different from implicit neural representations. The key distinction is that the reconstructed object is parameterized by explicit geometric primitives with learnable locations, covariances, and densities. 

\subsection{Simplex mesh representations}
\label{sec:simplex}

Simplex mesh representations are another form of explicit primitive representation, where the reconstruction domain is partitioned into irregular geometric cells such as triangles in two dimensions or tetrahedra in three dimensions. Instead of assigning values to a regular Cartesian voxel grid, the reconstructed object is represented over an adaptive mesh:
\begin{equation}
\mathcal{M} = (\mathcal{V}, \mathcal{E}, \mathcal{C}),
\end{equation}
where $\mathcal{V}$ denotes the vertices, $\mathcal{E}$ denotes the edges, and $\mathcal{C}$ denotes the simplex cells. The reconstructed signal can then be described as
\begin{equation}
\lambda(\mathbf{r}) \approx \sum_{k=1}^{K} \psi_k(\mathbf{r};\Theta_k),
\end{equation}
where each primitive corresponds to a simplex cell together with its geometric parameters and associated nodal or cell-wise values.

The main advantage of simplex mesh representations is their ability to allocate spatial resolution non-uniformly. This is useful when anatomical boundaries, lesions, or activity changes are spatially localized, because the mesh can be refined in important regions while remaining coarse elsewhere. A recent example is the anatomy-guided multi-resolution PET reconstruction method of Lesonen et al., which uses triangular meshes to balance reconstruction accuracy and computational cost \citep{lesonen2024anatomy}. This illustrates how mesh primitives can serve as an alternative to regular voxels when adaptive spatial discretization is desirable.

Compared with Gaussian primitives, simplex mesh representations are less common in recent AI-based radiological reconstruction. One reason is that irregular meshes require more specialized projection, interpolation, and optimization operators than regular voxel grids or Gaussian splatting pipelines.

\section{Implicit neural representations}
\label{sec7}

Unlike explicit representations that directly store signal values on grids, coefficients, or primitives, implicit neural representations model the target radiological image as a continuous coordinate-dependent function \citep{molaei2023implicit}. Let $\Omega \subset \mathbb{R}^{3}$ denote the imaging domain, and let $\lambda(\mathbf{r})$ denote the signal value at spatial location $\mathbf{r}\in\Omega$. In implicit reconstruction, the target image is parameterized as
\begin{equation}
\lambda_{\theta}:\Omega \rightarrow \mathbb{R}, \qquad
\mathbf{r} \mapsto \lambda_{\theta}(\mathbf{r}),
\label{eq:inr_general}
\end{equation}
where $\theta$ denotes the learnable parameters of the neural field. Once the model is optimized, the reconstructed image at any required resolution can be obtained by evaluating $\lambda_{\theta}(\mathbf{r})$ on the corresponding sampling coordinates.

Given the measurement $\mathbf{y}$ and the imaging forward operator $\mathcal{A}$, the reconstruction process can be written as
\begin{equation}
\theta^{*}=\arg\min_{\theta}\mathcal{D}\big(\mathcal{A}[\lambda_{\theta}],\mathbf{y}\big)+\mathcal{R}(\theta),
\label{eq:inr_recon}
\end{equation}
where $\mathcal{D}$ denotes the data fidelity term and $\mathcal{R}$ denotes the regularization term. Compared with explicit representations, Implicit neural representation avoids direct pixel or voxel storage and provides a continuous description of the target signal. In radiological imaging, current implicit methods can be more consistently divided into coordinate-based intensity field representations and radiance field representations. These representations are widely used in CT, MRI, PET, and US, as detailed in Table \ref{tbl7}.

{\scriptsize
\setlength{\tabcolsep}{3pt}
\renewcommand{\arraystretch}{1.05}
\begin{longtable}{
>{\raggedright\arraybackslash}p{0.14\linewidth}
>{\raggedright\arraybackslash}p{0.10\linewidth}
>{\raggedright\arraybackslash}p{0.25\linewidth}
>{\centering\arraybackslash}p{0.08\linewidth}
>{\raggedright\arraybackslash}p{0.33\linewidth}
}
\caption{Representative methods under the \textbf{Implicit neural representations} family. Coord. field and Radiance field correspond to the representations introduced in Section~\ref{sec7}.}
\label{tbl7}\\
\toprule
Rep. & Mod. & Paper & Year & Detail \\
\midrule

\multirow{11}{*}{Coord. field}
& \multirow{5}{*}{CT}
& Reed et al. \citep{reed2021dynamic}
& 2021
& An early medical INR method that reconstructs dynamic CT from limited views by combining a continuous neural field with parametric motion fields. \\
\cmidrule(l){3-5}

&
& NAF \citep{zha2022naf}
& 2022
& A neural attenuation field method for sparse-view CBCT reconstruction, representing attenuation coefficients as a continuous function of 3D coordinates. \\
\cmidrule(l){3-5}

&
& DIF-Net \citep{lin2023learning}
& 2023
& A deep intensity field method for extremely sparse-view CBCT reconstruction, learning a continuous coordinate-to-intensity representation from sparse projections. \\
\cmidrule(l){3-5}

&
& PINER \citep{10030867}
& 2023
& A prior-informed implicit neural representation method for test-time adaptation in sparse-view CT reconstruction. \\
\cmidrule(l){3-5}

&
& FACT \citep{shin2025sparse}
& 2025
& A fast sparse-view CBCT reconstruction method using a meta-learned neural attenuation field and hash-encoding regularization. \\
\cmidrule(l){2-5}

& \multirow{3}{*}{MRI}
& IMJENSE \citep{feng2023imjense}
& 2023
& A scan-specific INR framework for parallel MRI that jointly represents the image and coil sensitivity maps as continuous functions. \\
\cmidrule(l){3-5}

&
& NeSVoR \citep{xu2023nesvor}
& 2023
& An implicit neural representation method for slice-to-volume MRI reconstruction, recovering continuous fetal or neonatal MRI volumes from motion-corrupted slices. \\
\cmidrule(l){3-5}

&
& Spatiotemporal INR for dynamic MRI \citep{feng2025spatiotemporal}
& 2025
& A dynamic MRI reconstruction method that models the target sequence as a spatiotemporal coordinate field without requiring external training labels. \\
\cmidrule(l){2-5}

& PET
& IMREPET \citep{fan2025imrepet}
& 2025
& An unsupervised dynamic PET reconstruction method that parameterizes PET activity as a continuous spatiotemporal implicit neural representation. \\
\cmidrule(l){2-5}

& US
& ImplicitVol \citep{yeung2021implicitvol}
& 2021
& A sensorless 3D ultrasound reconstruction method using deep implicit representation to learn a coordinate-to-intensity ultrasound volume. \\

\cmidrule(l){2-5}
& CT/MRI
& NeRP \citep{shen2022nerp}
& 2022
& A prior-embedded INR framework that maps spatial coordinates to image intensity values and uses prior images to improve sparse CT and MRI reconstruction. \\

\midrule

\multirow{10}{*}{Radiance field}
& \multirow{5}{*}{CT}
& MedNeRF \citep{corona2022mednerf}
& 2022
& A medical neural radiance field framework that reconstructs 3D-aware CT projections from single-view or few-view X-rays. \\
\cmidrule(l){3-5}

&
& ACNeRF \citep{sun2024acnerf}
& 2024
& A medical NeRF method that improves single-view X-ray reconstruction by introducing alignment and pose correction for more accurate novel-view synthesis. \\
\cmidrule(l){3-5}

&
& UMedNeRF \citep{hu2024umednerf}
& 2024
& An uncertainty-aware medical NeRF method for single-view volumetric rendering and CT projection synthesis. \\
\cmidrule(l){3-5}

&
& VolumeNeRF \citep{Liu_VolumeNeRF_MICCAI2024}
& 2024
& A CT volume reconstruction method from a single projection view, integrating anatomical priors, projection attention, and Lambert--Beer law-based rendering. \\
\cmidrule(l){3-5}

&
& SAX-NeRF \citep{cai2024structure}
& 2024
& A structure-aware sparse-view X-ray 3D reconstruction method using radiance-field modeling with local and global ray sampling. \\
\cmidrule(l){2-5}

& \multirow{2}{*}{MRI}
& CuNeRF \citep{chen2023cunerf}
& 2023
& A cube-based NeRF framework for zero-shot arbitrary-scale medical image super-resolution and free-view synthesis in CT and MRI. \\
\cmidrule(l){3-5}

&
& Brain MRI NeRF reconstruction \citep{iddrisu20233d}
& 2023
& A NeRF-based method for reconstructing 3D-aware brain MRI representations from 2D MRI scans. \\
\cmidrule(l){2-5}

& \multirow{3}{*}{US}
& Ultra-NeRF \citep{wysocki2024ultra}
& 2024
& A physics-aware ultrasound NeRF that models attenuation, reflection, and scattering to synthesize view-dependent B-mode images from tracked 2D ultrasound scans. \\
\cmidrule(l){3-5}

&
& UlRe-NeRF \citep{guo2024ulre}
& 2024
& A 3D ultrasound neural rendering method that incorporates ultrasound reflection direction parameterization for improved view-dependent synthesis. \\
\cmidrule(l){3-5}

&
& NeRF-US \citep{dagli2024nerf}
& 2024
& A neural radiance field framework designed to reduce ultrasound imaging artifacts and improve 3D ultrasound reconstruction in real-world settings. \\

\bottomrule
\end{longtable}
}

\subsection{Coordinate-based intensity field representations}
\label{sec:coord_inr}

The most direct form of INR maps spatial coordinates to intensity, attenuation, or activity values:
\begin{equation}
\lambda_{\theta}(\mathbf{r}) = f_{\theta}\big(\gamma(\mathbf{r})\big),
\label{eq:coord_field}
\end{equation}
where $\gamma(\cdot)$ denotes an optional positional encoding. This formulation learns a continuous intensity field and is particularly suitable for sparse-view, limited-angle, and low-dose reconstruction, where continuity priors can compensate for missing measurements.

A practically important variant of this representation incorporates prior anatomical or functional information into the field model. This can be achieved by conditioning the implicit field on a prior embedding,
\begin{equation}
\lambda_{\theta}(\mathbf{r}) = f_{\theta}\big(\mathbf{r};\mathbf{z}\big), \qquad
\mathbf{z}=E(I_{\mathrm{prior}}),
\end{equation}
or by initializing the model parameters from a prior image,
\begin{equation}
\theta_{0}=\Psi(I_{\mathrm{prior}}),
\end{equation}
where $I_{\mathrm{prior}}$ denotes the prior image, and $E(\cdot)$ and $\Psi(\cdot)$ denote prior encoding functions.

Initially, this line of work appeared in PET imaging for reconstructing normal-dose images from low-dose scans. Gong et al. integrated the deep image prior framework with a nonlocal operation, using a kernel matrix layer for feature denoising and embedding linear kinetic models as convolutional layers \citep{gong2021direct}. Experimental results demonstrated better performance compared with traditional methods and kernel-based approaches \citep{gong2021direct}. This formulation improves parametric image reconstruction by enhancing image quality and suppressing noise.

Subsequently, NeRP \citep{shen2022nerp} extended this idea to multi-modality imaging and established a representative coordinate-based INR framework for medical image reconstruction. Unlike explicit methods, NeRP employs an MLP to directly map spatial coordinates to intensity values, thereby learning a continuous representation of the target image without explicit pixel storage. At the same time, it incorporates prior image information into the neural representation, enabling the model to extract structural cues and optimize the reconstruction process jointly with physical measurements. NeRP achieved a PSNR of 39.06 dB and an SSIM of 0.986 using only 20 projections, outperforming methods such as FBP and GRFF \citep{shen2022nerp}.

Numerous subsequent studies \citep{shi2024implicit,liu2024geometry} have shown that this type of coordinate-based INR can produce high-quality radiological image reconstruction, particularly in tasks involving low-dose to normal-dose recovery. However, it also has clear limitations. First, many prior-guided formulations require each subject to have at least one normal-dose scan, which restricts practical applicability \citep{shen2022nerp}. Second, in real-world scenarios, prior images may not always be available or well aligned because of anatomical variation, patient motion, or scanner differences \citep{adlam2020exploring}. These factors limit the robustness of prior-guided coordinate-based INR in routine clinical use.

\subsection{Radiance field representations}
\label{sec:nerf}

Radiance field representation is a rendering-oriented subtype of INR. Instead of predicting only a scalar intensity field, it jointly models density and view-dependent appearance:
\begin{equation}
F_{\theta}(\mathbf{r},\mathbf{d})=
\big(\sigma(\mathbf{r}),\mathbf{c}(\mathbf{r},\mathbf{d})\big),
\label{eq:nerf_field}
\end{equation}
where $\sigma(\mathbf{r})$ denotes density or attenuation-related quantity and $\mathbf{c}(\mathbf{r},\mathbf{d})$ denotes the view-dependent signal associated with viewing direction $\mathbf{d}$. The final image is synthesized through differentiable volume rendering:
\begin{equation}
\hat I(\mathbf{u})=\int T(t)\,\sigma\big(\mathbf{r}(t)\big)\,
\mathbf{c}\big(\mathbf{r}(t),\mathbf{d}\big)\,dt,
\label{eq:nerf_render}
\end{equation}
where $\mathbf{r}(t)$ parameterizes the ray corresponding to detector coordinate $\mathbf{u}$ and $T(t)$ denotes the accumulated transmittance.

Neural Radiance Fields have emerged as a powerful technique for 3D reconstruction and rendering. Unlike traditional methods that depend on discrete volume representations, NeRF offers enhanced performance in generating detailed medical images \citep{mildenhall2021nerf,Liu_VolumeNeRF_MICCAI2024}, especially in CT-related tasks.

MedNeRF \citep{corona2022mednerf} was the first to apply NeRF to radiological imaging. It employs self-supervised learning and extends the Generative Radiance Field framework with a GAN architecture tailored to the medical domain. This approach combines self-supervised learning and data augmentation to enhance reconstruction fidelity from sparse data \citep{corona2022mednerf}. Trained and validated on digitally reconstructed radiographs, MedNeRF demonstrated promising rendering quality when reconstructing from single-view X-rays. However, its heavy dependence on single-view input also leads to obvious artifacts, which limits clinical applicability.

VolumeNeRF \citep{Liu_VolumeNeRF_MICCAI2024} further improves this line by employing a 3D encoder-decoder architecture to reconstruct CT volumes from 2D X-rays. It incorporates anatomical priors from likelihood and average CT images, introduces a projection attention module to enhance spatial alignment, and uses Lambert--Beer law-based volumetric rendering for additional supervision. This method improves reconstruction quality and preserves anatomical edges and details more effectively.

When applied to ultrasound imaging, Ultra-NeRF \citep{wysocki2024ultra} provides another representative example. It is an implicit neural representation model for ultrasound imaging that synthesizes view-dependent B-mode images from multiple 2D ultrasound scans. By incorporating physics-based rendering formulas and ray tracing, Ultra-NeRF explicitly models attenuation, reflection, and scattering, thereby capturing acoustic shadowing and occlusion effects that are difficult to represent with conventional volumetric synthesis.

Following this, Chen et al. \citep{chen2023cunerf} introduced CuNeRF, a zero-shot framework for CT and MRI reconstruction. It reconstructs high-resolution medical volumes from low-resolution inputs without paired high-resolution data, using cubic grids for sampling, isotropic volume rendering, and hierarchical adaptive sampling. CuNeRF outperformed state-of-the-art methods while enabling free-viewpoint synthesis across modalities.

Although radiance field representations have attracted widespread attention in recent years, their computational complexity remains a major obstacle to clinical deployment \citep{molaei2023implicit}. These methods usually require learning a neural field for each signal individually, which results in substantial memory usage and long optimization time, especially for high-dimensional medical volumes. In addition, under sparse sampling conditions, radiance field reconstruction may still suffer from blurring and structural distortion, which is particularly critical in medical imaging.

\section{Discussions and Challenges}
\label{sec8}

\subsection{Comparison of Representation Families}
\label{sec:comparison}

Building upon the above discussion, we organize AI-based 3D reconstruction in radiological imaging into four representation families: discrete grid representations, explicit basis expansion representations, explicit primitive representations, and implicit neural representations. These families differ not only in reconstruction quality and computational efficiency, but also in adaptability, simplicity, and interpretability. To keep the comparison fully consistent with the taxonomy used throughout this review, Table~\ref{tbl8} summarizes these four representation families. For reference, traditional iterative reconstruction is also included as a non-AI baseline.

We evaluate these paradigms according to five criteria: \textbf{effectiveness} (Effctv.), \textbf{efficiency} (Effcy.), \textbf{adaptability} (Adapt.), \textbf{simplicity} (Simpl.), and \textbf{interpretability} (Intrpt.). The scores in Table~\ref{tbl8} are intended as an ordinal qualitative summary of the reviewed literature rather than a strict quantitative meta-analysis, because the reported studies differ substantially in imaging modality, anatomy, acquisition protocol, sparsity level, and evaluation metrics. A score of 1 indicates relatively weak performance, 3 indicates moderate or task-dependent performance, and 5 indicates relatively strong performance within the scope of the reviewed literature. The scores were assigned by jointly considering the representative methods summarized in Sections~\ref{sec4}--\ref{sec7}, their reported image quality and computational characteristics, and the representation-level properties discussed throughout this review. To improve transparency, the scoring rubric used for Table~\ref{tbl8} is summarized in Table~\ref{tbl:score_rubric}.

\begin{table}[t]
\caption{Rubric used for assigning the ordinal scores in Table~\ref{tbl8}. Scores 2 and 4 indicate intermediate levels between neighboring anchors.}
\label{tbl:score_rubric}
\centering
\footnotesize
\begin{tabularx}{\linewidth}{p{0.16\linewidth}|X|X|X}
\hline
\hline
Criterion & Score 1 & Score 3 & Score 5 \\
\hline
Effectiveness &
Limited reconstruction fidelity under challenging settings; prone to artifacts, noise amplification, or loss of anatomical structure. &
Competitive performance in standard settings, but quality may degrade under severe sparse-view, low-dose, undersampled, or cross-domain conditions. &
Consistently strong reconstruction quality in challenging settings, with good preservation of anatomical structure, spatial continuity, and clinically relevant details. \\
\hline
Efficiency &
High computational cost, slow optimization or rendering, or large memory consumption that limits practical deployment. &
Moderate computational cost; feasible for offline reconstruction but still limited for real-time or large-scale clinical use. &
Fast reconstruction or rendering, good parallelization, and relatively low memory or parameter burden, supporting practical deployment. \\
\hline
Adaptability &
Narrowly designed for specific imaging geometries, modalities, or acquisition conditions; requires substantial redesign when transferred. &
Applicable to several related tasks or modalities, but usually requires task-specific tuning, priors, or acquisition-dependent modifications. &
Broadly compatible with multiple imaging modalities, reconstruction tasks, and clinical pipelines with limited representation-level modification. \\
\hline
Simplicity &
Complex parameterization, multi-stage optimization, task-specific initialization, or difficult implementation. &
Moderate implementation complexity; some specialized operators, priors, or training strategies are required. &
Straightforward formulation and implementation; easy integration with conventional image reconstruction or post-processing pipelines. \\
\hline
Interpretability &
Representation parameters are abstract and difficult to connect with physical imaging models or anatomical meaning. &
Partially interpretable; some components correspond to images, coordinates, bases, or primitives, but the learned mapping remains partly opaque. &
Highly traceable formulation, with parameters or update steps closely related to physical models, image grids, coefficients, or explicit reconstruction operations. \\
\hline
\hline
\end{tabularx}
\normalsize
\end{table}

\begin{table}[h]
\caption{Comparison of the four representation families for radiological image reconstruction. Traditional iterative reconstruction is included only as a reference baseline.}
\label{tbl8}
\centering
\footnotesize
\begin{tabular}{p{0.46\linewidth}|c|c|c|c|c}
\hline
\hline
Method / Family & Effctv. & Effcy. & Adapt. & Simpl. & Intrpt. \\
\hline
Traditional iterative reconstruction (reference) & 3 & 2 & 4 & 3 & 5 \\
\hline
Discrete grid representation             & 4 & 4 & 5 & 5 & 3 \\
Explicit basis expansion representation  & 3 & 3 & 4 & 3 & 4 \\
Explicit primitive representation        & 4 & 5 & 3 & 2 & 3 \\
Implicit neural representation           & 5 & 2 & 3 & 2 & 2 \\
\hline
\hline
\end{tabular}
\normalsize
\end{table}

As shown in Table~\ref{tbl8}, traditional iterative reconstruction remains the most interpretable paradigm because its formulation is directly linked to imaging physics, statistical assumptions, and optimization objectives. However, its efficiency is often limited by repeated forward and backward projection operations, and its effectiveness under highly sparse, low-dose, or noisy conditions is usually task-dependent.

Among AI-based methods, discrete grid representations remain the most intuitive and adaptable solution, because they are naturally compatible with conventional medical imaging pipelines and can be implemented with relatively simple architectures. They generally achieve strong reconstruction performance in CT, MRI, PET, and SPECT tasks while maintaining favorable efficiency and ease of deployment. However, their fixed sampling structure may limit fine spatial continuity, especially in slice-wise settings.

Explicit basis expansion representations provide a broader explicit mathematical framework than regular grids. By representing the target image through basis coefficients, they can better capture smooth local image structure while retaining a relatively interpretable coefficient-based form. Nevertheless, their practical performance depends strongly on the selected basis functions, kernel design, and conditioning of the imaging operator, which leads to moderate scores in effectiveness, efficiency, and simplicity.

Explicit primitive representations, especially Gaussian primitive methods, achieve a favorable balance between reconstruction quality and rendering efficiency. Their adaptive spatial support makes them attractive for sparse-view reconstruction, view synthesis, and real-time rendering scenarios. However, compared with grid-based methods, they usually require more specialized initialization, rendering, and optimization strategies, which reduces their simplicity and limits their adaptability across tasks.

Implicit neural representations provide the strongest ability to model continuous structures and often achieve high reconstruction quality under sparse-view or low-dose settings. However, they are generally more difficult to optimize and interpret than explicit representations. In particular, radiance field methods, as a rendering-oriented subtype of implicit neural representation, usually involve higher computational cost, whereas coordinate-based intensity field methods are relatively simpler but still less transparent than explicit formulations.

Overall, explicit methods are generally easier to integrate into existing medical imaging workflows and are often more efficient at inference time, whereas implicit methods are more flexible in modeling continuous structures and complex priors. At the same time, nearly all AI-based reconstruction approaches still face interpretability challenges when compared with traditional iterative reconstruction.

\subsection{Imaging Quality Assessment}

As AI-driven radiological reconstruction evolves from discrete grid representations in Section~\ref{sec4} and explicit basis expansion representations in Section~\ref{sec5} to explicit primitive representations in Section~\ref{sec6} and implicit neural representations in Section~\ref{sec7}, commonly used image quality metrics such as PSNR and SSIM continue to improve \citep{wang2004image}. However, medical image quality assessment must go beyond computational benchmarks and consider clinical relevance \citep{chow2016review}, including artifact severity \citep{wuest2015improved}, registration accuracy \citep{ketcha2017effects}, anatomical fidelity, and downstream diagnostic utility \citep{yang2023ct}. Therefore, clinician evaluation remains indispensable for complementing quantitative metrics.

In the natural image domain, numerous studies have shown that combining Image Quality Assessment (IQA) \citep{athar2019comprehensive} with Image Quality Metrics (IQM) \citep{ahumada1993computational} provides a more effective means of evaluating perceptual quality. This strategy has been widely adopted in restoration tasks such as deblurring \citep{da2023image} and low-light enhancement \citep{kandula2023illumination,li2024lime}. Similar efforts have also appeared in medical imaging \citep{kastryulin2023image,leveque2021comparative,mirza2023comparative}. For radiological reconstruction, future evaluation protocols should therefore combine pixel-level fidelity, perceptual similarity, structural consistency, physical plausibility, and reader studies, rather than relying on a single metric.

At the same time, the adoption of more diverse and task-aware evaluation criteria can further improve the reliability of reconstruction assessment. For radiological reconstruction, future benchmarks should place greater emphasis on clinical usability, anatomical consistency, and downstream diagnostic value.

\subsection{Efficiency and Effectiveness}

Early CNN-based reconstruction methods enabled fast but relatively basic image recovery in discrete grid representations. In contrast, more advanced models such as diffusion-based volumetric methods and implicit neural representations often face a pronounced trade-off between reconstruction quality and computational speed. Explicit primitive representations, especially Gaussian-based methods, partly alleviate this issue by combining adaptive local modeling with efficient parallel rendering. However, jointly achieving high reconstruction fidelity, low latency, and low computational cost remains a central challenge in medical image reconstruction \citep{zhou2021review,prevedello2019challenges}.

Transfer learning has been widely adopted to reduce training time and computational cost by pretraining models on large-scale datasets and fine-tuning them for specific tasks \citep{zhuang2020comprehensive}. In the medical domain, it has proven effective in classification \citep{kim2022transfer}, segmentation \citep{karimi2021transfer}, and related tasks \citep{raghu2019transfusion}. The emergence of the Segment Anything Model \citep{kirillov2023segment} has further shown that large pretrained models can be adapted to medical imaging through fine-tuning \citep{ma2024segment}. However, unlike classification and segmentation, radiological reconstruction is tightly coupled with imaging physics and acquisition geometry, which makes direct reuse of generic pretrained models more difficult \citep{zhang2017deep,knoll2019assessment}. Imaging-specific priors therefore remain essential.

Edge computing is another promising strategy, because it moves computationally intensive tasks to edge devices or cloud platforms and enables distributed processing \citep{chen2019deep}. This strategy has already been explored in radiological image reconstruction and computer-aided detection \citep{isosalo2023local,zhang20213d,zhang2025collaborative}. Meanwhile, although low-dose and sparse-view reconstruction reduce acquisition burden, the computational cost of large AI models has become increasingly prominent. To reduce this cost, lightweight end-to-end architectures and knowledge distillation have attracted growing attention \citep{gou2021knowledge}. In reconstruction-related medical tasks, distillation has shown potential for reducing inference cost while preserving performance \citep{zhou2024distillation,wu2024adaptive}.

\subsection{Privacy Protection}

Given the sensitivity of medical imaging data, privacy-preserving AI-based reconstruction is of critical importance \citep{williamson2024balancing}. A first line of work focuses on reconstruction strategies that reduce or avoid reliance on large centralized training datasets. Examples include unsupervised models based on Gaussian mixture modeling \citep{li2022noise}, unbiased risk estimation \citep{aggarwal2022ensure}, or 3DGS-based architectures \citep{cai2024structureawaresparseviewxray3d,cai2025radiative}. Since these methods can operate without direct access to large collections of patient data, they offer a promising path toward stronger privacy protection \citep{dike2018unsupervised}.

Federated learning provides another important solution by training models on decentralized data sources without moving sensitive patient data across institutions \citep{guan2024federated}. This idea has already been applied to MRI reconstruction and has shown performance close to that of multi-institutional centralized training \citep{feng2022specificity,elmas2022federated}. In addition, domain adaptation \citep{guan2021domain,han2018deep}, domain generalization \citep{yoon2024domain}, and test-time training \citep{zhao2024test,he2021autoencoder} can reduce dependence on direct access to private clinical datasets by improving robustness across imaging domains.

Further protection can be achieved through differential privacy, which constrains the leakage of individual data points during model training \citep{ziller2021medical}. Homomorphic encryption enables computation on encrypted medical data without direct disclosure of sensitive information \citep{yang2019secure,dutil2021application}. Blockchain-based secure data sharing has also been explored for transparent and privacy-preserving medical data exchange \citep{xi2022review,shen2019privacy}. These techniques suggest that future radiological reconstruction systems should jointly consider image quality, computational efficiency, data availability, and privacy protection.

\subsection{Model Interpretability}

The limited transparency of complex AI-based reconstruction models remains a major obstacle to clinical trust and adoption \citep{eke2024role}. To improve interpretability, recent studies attempt to decompose reconstruction into more traceable steps or to align learning modules with physically meaningful operations \citep{salahuddin2022transparency}. Existing strategies can generally be divided into ante-hoc interpretability \citep{li2018tell} and post-hoc interpretability \citep{selvaraju2017grad,sundararajan2017axiomatic}.

In radiological image reconstruction, Zhang et al. proposed DUN-SA by unfolding iterative optimization stages into network modules with clear physical meanings \citep{zhang2025deep}. This design improves interpretability and addresses the lack of explainability in deep learning-based multimodal MRI reconstruction. More broadly, explicit representations are often easier to interpret because their parameters are directly associated with grids, coefficients, or primitives, whereas implicit neural fields tend to be more abstract and harder to explain. Improving interpretability will therefore be essential for the safe and effective clinical deployment of AI-based radiological reconstruction.

\subsection{Future Research Directions}
\label{sec:future-directions}

The above challenges indicate several important directions for future research.

First, representation design should be more tightly integrated with imaging physics. Existing explicit and implicit representations each have distinct advantages. Discrete grid representations and explicit basis expansion representations are compatible with conventional reconstruction pipelines, explicit primitive representations provide efficient adaptive modeling, and implicit neural representations offer continuous signal modeling. A promising direction is to develop hybrid representations that combine these strengths while incorporating physically accurate forward models, acquisition geometry, uncertainty estimation, and modality-specific constraints.

Second, clinically meaningful evaluation protocols are needed. Current studies often report PSNR, SSIM, or related image-level metrics, but these metrics do not fully reflect diagnostic reliability. Future benchmarks should include standardized multi-center datasets, realistic low-dose or sparse-view protocols, task-aware quantitative metrics, reader studies, and downstream clinical tasks such as lesion detection, segmentation, registration, treatment planning, or disease classification. Such evaluation would make comparisons across representation families more rigorous and clinically relevant.

Third, efficient and deployable reconstruction models should be prioritized. Although large generative models and implicit neural representations can improve reconstruction quality, their computational burden may restrict clinical use. Future work should explore lightweight architectures, knowledge distillation, hardware-aware optimization, edge-cloud collaboration, and fast differentiable rendering. These strategies are especially important for time-sensitive scenarios such as interventional imaging, dynamic imaging, and real-time ultrasound or cone-beam CT reconstruction.

Fourth, generalizable and privacy-preserving learning remains essential. Medical imaging data are heterogeneous across scanners, institutions, protocols, and patient populations, while direct data sharing is often restricted. Future reconstruction systems should combine self-supervised or unsupervised learning, domain adaptation, domain generalization, test-time adaptation, federated learning, differential privacy, and secure computation. This direction is critical for building robust models that can perform reliably across clinical environments without compromising patient privacy.

Fifth, interpretability, reliability, and uncertainty awareness should be incorporated into reconstruction pipelines. For clinical deployment, models should not only produce visually plausible images but also provide traceable reconstruction mechanisms, confidence estimates, and failure warnings. Physics-guided network design, unfolded optimization, interpretable primitive parameters, uncertainty quantification, and causal analysis may help clinicians understand when AI-generated reconstructions can be trusted.

Taken together, future progress in AI-based 3D radiological reconstruction will likely depend on the coordinated development of representation learning, imaging physics, efficient computation, privacy-preserving training, and clinically grounded validation.

\section{Conclusion}
\label{sec9}

Artificial intelligence has profoundly advanced 3D reconstruction for radiological imaging across CT, MRI, PET, SPECT, and ultrasound. In recent years, the field has gradually evolved from an architecture-centered perspective toward a representation-centered understanding of reconstruction methods.

In this review, we organized existing studies into four representation families: discrete grid representations, explicit basis expansion representations, explicit primitive representations, and implicit neural representations. This taxonomy provides a clearer view of how reconstruction targets are parameterized in modern radiological imaging. In particular, Gaussian-based methods belong to explicit primitive representations, while radiance field methods should be regarded as a subtype of implicit neural representation rather than an independent top-level category.

Based on this perspective, we summarized representative methods, benchmark datasets, evaluation metrics, and the main challenges in the field. Our analysis suggests that no single representation family is universally optimal. Discrete grid representations remain strong in simplicity and adaptability, basis expansion representations provide a broader explicit mathematical framework, primitive representations offer efficient adaptive modeling, and implicit neural representations are advantageous for continuous high-fidelity modeling.

Overall, future progress in AI-based radiological reconstruction will depend on better integration of representation design with imaging physics, more clinically meaningful evaluation, higher computational efficiency, stronger privacy protection, and improved interpretability. We expect that continued advances along these directions will further promote the development of low-cost, high-quality, and clinically practical radiological image reconstruction systems.
\backmatter

\bmhead{Funding}
This work was supported by the National Natural Science Foundation of China (Grant No. 62306003) and the Open Research Fund of Guangdong Laboratory of Artificial Intelligence and Digital Economy (SZ) (Grant No. GML-KF-24-29).

\bmhead{Author contributions}
Yuezhe Yang: Conceptualization of this study, Methodology, Literature investigation and analysis, Writing - Original Draft, Writing - Review \& Editing, Visualization. Lei Bi: Methodology, Writing - Review \& Editing, Project administration. Boyu Yang: Literature investigation and analysis, Visualization, figures and tables design, Software. Yaqian Wang: Literature investigation and analysis, Writing - Review \& Editing. Yang He: Literature investigation and analysis, Visualization, figures and tables design. Yige Peng: Writing - Review \& Editing, Supervision. Zhe Jin: Writing - Review \& Editing, Supervision. Xingbo Dong: Methodology, Writing - Review \& Editing, Project administration. Jinman Kim: Methodology, Writing - Review \& Editing, Project administration.

\bmhead{Data availability}
The supporting materials for this review are available at \url{https://github.com/Bean-Young/AI4Radiology}.

\section*{Declarations}

\bmhead{Competing interests}
The authors declare no competing interests.

\bmhead{Ethical approval}
This article does not contain any studies with human participants or animals performed by any of the authors.

\bibliography{cas-refs}

\end{document}